\def\year{2022}\relax
\documentclass[letterpaper]{article} 
\usepackage{aaai22}  
\usepackage{times}  
\usepackage{helvet}  
\usepackage{courier}  
\usepackage[hyphens]{url}  
\usepackage{graphicx} 
\usepackage{natbib}  
\usepackage{caption} 
\DeclareCaptionStyle{ruled}{labelfont=normalfont,labelsep=colon,strut=off} 
\usepackage{algorithm}
\usepackage{algorithmic}
\usepackage{amsmath,amssymb,amsfonts}
\usepackage{textcomp}
\usepackage{subcaption}
\usepackage{mathtools}
\usepackage{booktabs,siunitx}
\usepackage{makecell}
\usepackage{multirow}
\usepackage{xcolor}

\usepackage{newfloat}
\usepackage{listings}
\floatstyle{ruled}
\newfloat{listing}{tb}{lst}{}
\floatname{listing}{Listing}
\title{ErfAct and Pserf: Non-monotonic Smooth Trainable Activation Functions}
\author {
    Koushik Biswas\textsuperscript{\rm 1},
    Sandeep Kumar \textsuperscript{\rm 1, 2},
    Shilpak Banerjee\textsuperscript{\rm 3}, Ashish Kumar Pandey\textsuperscript{\rm 3}
}
\affiliations {
    \textsuperscript{\rm 1} Department of Computer Science, IIIT Delhi, New Delhi, India 110020\\
    \textsuperscript{\rm 2}Department of Mathematics, Shaheed Bhagat Singh College, University of Delhi, New Delhi, India 110017\\
    \textsuperscript{\rm 3}Department of Mathematics, IIIT Delhi, New Delhi, India 110020\\
    koushikb@iiitd.ac.in,
    sandeep\_kumar@sbs.du.ac.in,
    shilpak@iiitd.ac.in,
    ashish.pandey@iiitd.ac.in
}

\usepackage{bibentry}

\begin{document}

\maketitle

\begin{abstract}
An activation function is a crucial component of a neural network that introduces non-linearity in the network. The state-of-the-art performance of a neural network depends also on the perfect choice of an activation function. We propose two novel non-monotonic smooth trainable activation functions, called ErfAct and Pserf. Experiments suggest that the proposed functions improve the network performance significantly compared to the widely used activations like ReLU, Swish, and Mish. Replacing ReLU by ErfAct and Pserf, we have $5.68\%$ and $5.42\%$ improvement for top-1 accuracy on Shufflenet V2 (2.0x) network in CIFAR100 dataset, $2.11\%$ and $1.96\%$ improvement for top-1 accuracy on Shufflenet V2 (2.0x) network in CIFAR10 dataset, $1.0\%$, and $1.0\%$ improvement on mean average precision (mAP) on SSD300 model in Pascal VOC dataset.
\end{abstract}
\section{Introduction}
The choice of activation function in a deep learning architecture can have a significant impact on the training and performance of the neural network. The machine learning community has so far relied on hand-designed activations like ReLU \cite{relu}, Leaky ReLU \cite{lrelu} or their variants. ReLU, in particular, remains widely popular due to faster training times and decent performance.  However, evidence suggests that considerable gains can be made when more sophisticated activation functions are used to design networks. For example, activation functions such as ELU \cite{elu}, Parametric ReLU (PReLU) \cite{prelu}, ReLU6 \cite{relu6}, PAU \cite{pau}, OPAU \cite{opau}, ACON \cite{acon}, Mish \cite{mish}, GELU \cite{gelu}, Swish \cite{swish}, Serf \cite{serf} etc. have appeared as powerful contenders to the traditional ones. Though ReLU remains a go-to choice in both research and practice, it has certain well-documented shortcomings such as non-zero mean \cite{elu}, non-differentiability and negative missing, which leads to the infamous vanishing gradients problem (also known as the dying ReLU problem). Worth noting that prior to the introduction of ReLU, Tanh and Sigmoid were popularly used, but performance gains and training time gains achieved by ReLU led to their decline. 


\section{Related works and motivation}

The newer activation functions are obtained by combining well-known functions with simple forms in 
various ways, often using hyper-parameters or trainable parameters. In the case of trainable parameters, we optimize them during the training process itself, yielding networks that are better fitted. In the case of trainable parameters, note that the actual activation function curve may change in different layers during backpropagation. For example, SiLU \cite{silu} shows good performance over known activation functions while Swish \cite{swish} is a trainable version of SiLU, which is a non-linear, non-monotonic activation function. Among the activation mentioned in the previous section, Swish, PReLU, PAU, ACON, and OPAU are trainable activation functions. Swish is a non-monotonic activation function and shows promise across a variety of deep learning tasks. Mish is one of the popular functions proposed recently and gained popularity due to its effectiveness in object detection task on COCO dataset \cite{coco} in Yolo \cite{yolo} models. GELU is very similar to Swish and gained attention due to its effectiveness in computer vision and natural language processing tasks. It is also used in popular architectures like GPT-2 \cite{gpt2} and GPT-3 \cite{gpt3}. Apart from using a combination of known functions, a somewhat fundamentally different technique to construct activation functions is to use perturbation or approximations to well-known activation functions to remove some of the shortcomings yet retain the positive aspects. Recent successful examples where this strategy was employed includes PAU and OPAU, which are activations based on approximation of Leaky ReLU by rational polynomials were constructed. 

Motivated from these works, we have proposed two activation functions with trainable parameters, we call them ErfAct and Pserf and shown that they are more effective than conventional activation functions like ReLU, Leaky ReLU, PReLU, ReLU6, Swish, Mish or GELU in a wide range of standard deep learning problems. We summarize the paper as follows: 

\begin{figure*}[!t]
    \begin{minipage}[t]{.32\linewidth}
        \centering
    
        \includegraphics[width=\linewidth]{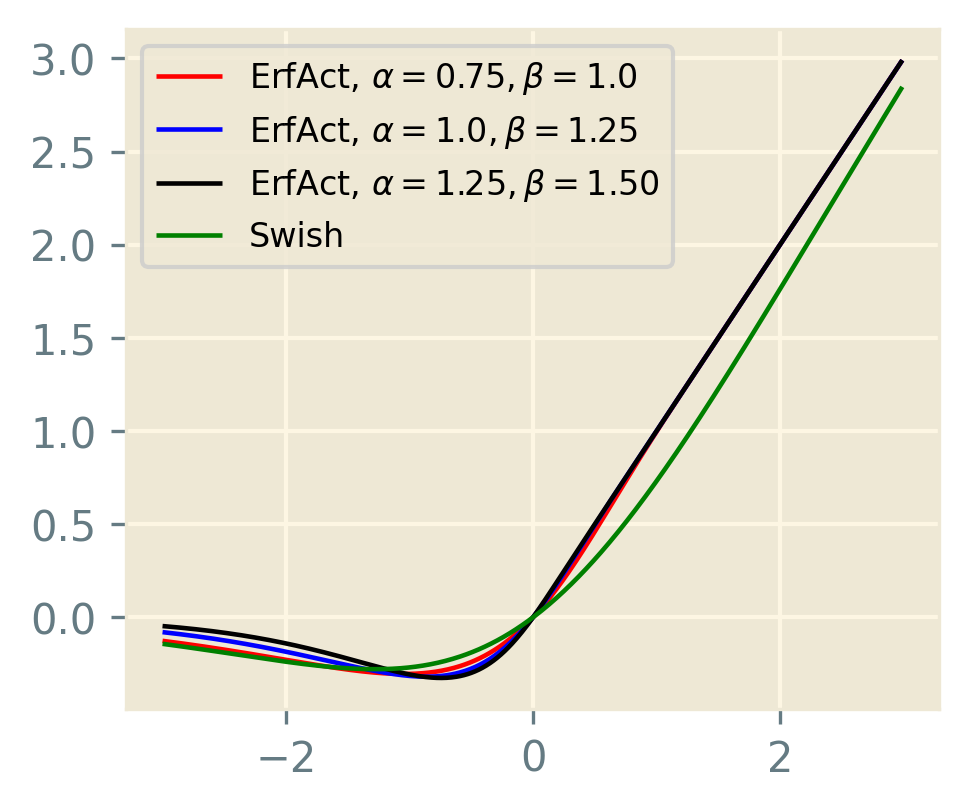}
        
        \caption{Swish and ErfAct activation for different values of $\alpha$ and $\beta$}
        \label{erf1}
    \end{minipage}
   \hfill
    \begin{minipage}[t]{.32\linewidth}
        \centering
        
       \includegraphics[width=\linewidth]{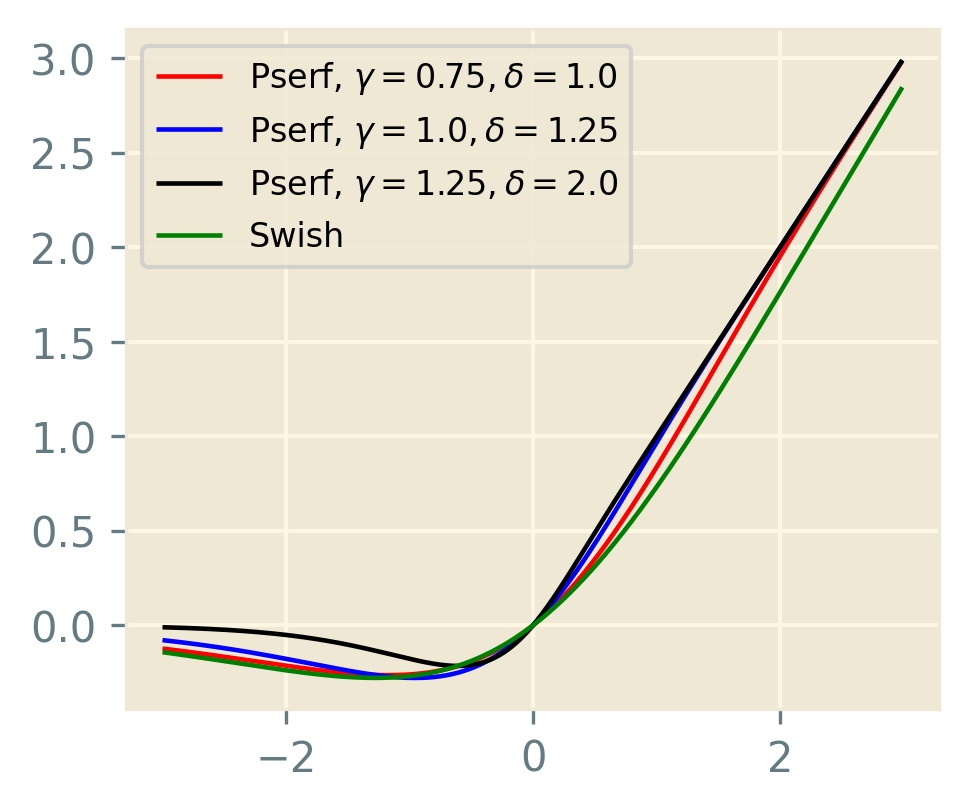}
      
        \caption{Swish and Pserf activation for different values of $\gamma$ and $\delta$}
        \label{erf2}
    \end{minipage}  
    \hfill
    \begin{minipage}[t]{.32\linewidth}
        \centering
        
       \includegraphics[width=\linewidth]{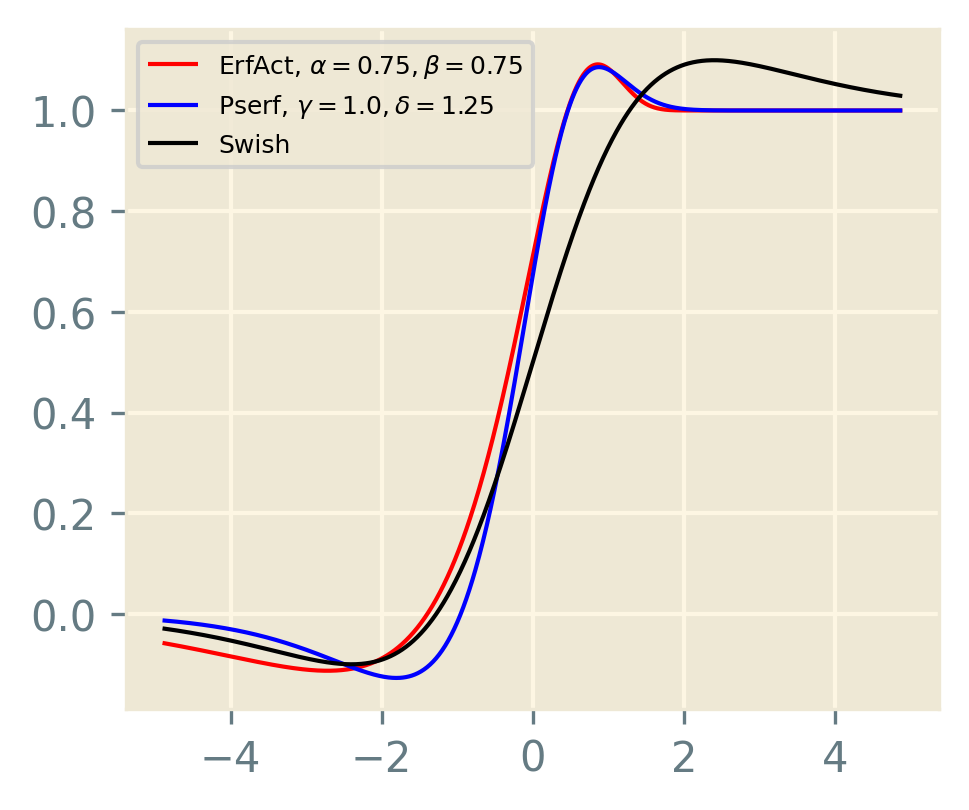}
     
        \caption{First order derivative of ErfAct, Pserf, and Swish}
        \label{der}
    \end{minipage}  
\end{figure*}

\begin{itemize}
    \item We have proposed two new novel trainable activation functions, which are the smooth approximation of ReLU.
    \item In a wide range of deep learning tasks, the proposed functions outperform widely used activation functions. 
\end{itemize}
\section{ErfAct and Pserf}
As motivated earlier, in this paper, we present, ErfAct and Parametric-Serf (Pserf), two novel trainable activation functions which outperforms the widely used activations and has the potential to replace them. ErfAct and Pserf is defined as
\begin{align}
\text{ErfAct}: \mathcal{F}_1(x;\alpha, \beta):= & x \ \text{erf}(\alpha e^{\beta x}),\\
\text{Pserf}:\ 
\mathcal{F}_2(x;\gamma, \delta):=& x \ \text{erf}(\gamma ln(1+e^{\delta x}))
\end{align}
where $\alpha, \beta, \gamma,$ and $\delta$ are trainable parameters (they can be used as hyper-parameters as well) and `erf' is the error function also known as the Gauss error function and defined as 
\begin{align}
    \text{erf}(x) = \dfrac{2}{\sqrt{\pi}} \int_{0}^{x} e^{-t^2} \,dt 
\end{align}
The corresponding derivatives of the proposed activations are

\begin{align}
\frac{d}{dx} \mathcal{F}_1(x;\alpha, \beta)=& \  \text{erf}(\alpha e^{\beta x}) + \dfrac{2x\alpha \beta}{\sqrt{\pi}} e^{\beta x} e^{-({{\alpha e^{\beta x}}})
^2}
\end{align}
\begin{align}
\frac{d}{dx} \mathcal{F}_2(x;\gamma, \delta)=& \text{erf}(\gamma ln(1+e^{\delta x}))\notag \\ + & \dfrac{2x\gamma \delta}{\sqrt{\pi}} \dfrac{e^{\delta x}}{1+e^{\delta x}} e^{-(\gamma ln(1+e^{\delta x}))^2}
\end{align}
where
\begin{align}
\frac{d}{dx}
    \text{erf}(x) = & \ \frac{2}{\sqrt{\pi}}e^{-x^2}
\end{align}

\hspace*{-0.35cm}ErfAct and Pserf are non-monotonic, zero-centered, continuously differentiable, unbounded above but bounded below, and trainable functions. Figures~\ref{erf1} and ~\ref{erf2} show the plots for $\mathcal{F}_1(x;\alpha, \beta)$ and $\mathcal{F}_2(x;\gamma, \delta)$ activation functions for different values of $\alpha$, $\beta$, and $\gamma, \delta$ respectively. A comparison between the first derivative of $\mathcal{F}_1(x;\alpha, \beta)$, $\mathcal{F}_2(x;\gamma, \delta)$, and Swish are given in Figures~\ref{der},  different values of $\alpha$, $\beta$, and $\gamma, \delta$ respectively. From the figures~\ref{erf1} and ~\ref{erf2} it is evident that the parameters $\alpha$, $\beta$, and $\gamma, \delta$ controls the slope of the curves for the proposed activations in both positive and negative axis. The proposed functions converges to some known functions for specific values of the parameters. For example, $\mathcal{F}_1(x;0, \beta)$, $\mathcal{F}_2(x;0, \delta)$ are zero function while $\mathcal{F}_1(x;\alpha, 0)$, $\mathcal{F}_2(x;\gamma, 0)$ are linear functions. In particular, $\mathcal{F}_2(x;1, 1)$ share the equivalent form as Serf\cite{serf} which is a non-parametric form of Pserf. Also, The proposed functions can be seen as a smooth approximation of ReLU. 
\begin{align*}
\begin{split}
    \lim_{\beta\to\infty}\mathcal{F}
    _1(x;\alpha, \beta)={}& \operatorname{ReLU}(x),\\
& \forall x\in\mathbb{R}\text{ for any fixed }\alpha>0.
\end{split}
\end{align*}
\begin{align*}
    \begin{split}
    \lim_{\delta\to\infty}\mathcal{F}
    _2(x;\gamma, \delta)={}& \operatorname{ReLU}(x),\\
& \forall x\in\mathbb{R}\text{ for any fixed }\gamma>0.
\end{split}
\end{align*}

\hspace*{-0.45cm} For any $K$, a compact (closed and bounded) subset of $\mathbb{R}^n$, the set of neural networks with ErfAct (or Pserf) activation functions is dense in $C(K)$, the space of all continuous functions over $K$ (see \cite{pau}). This follows from the next proposition, as the proposed activation functions are not polynomials. 

\smallskip

\noindent\textbf{Proposition (Theorem 1.1 in Kidger and Lyons, 2019 \cite{universal}) :-} Let $\rho: \mathbb{R}\rightarrow \mathbb{R}$ be any continuous
function. Let $N_n^{\rho}$ represent the class of neural networks with activation function $\rho$, with $n$ neurons in the input layer, one neuron in the output layer, and one hidden layer with an arbitrary number of neurons. Let $K \subseteq \mathbb{R}^n$
be compact. Then $N_n^{\rho}$ is dense in $C(K)$ if and only if $\rho$ is non-polynomial.

\section{Experiments}
We have compared our proposed activations against ten popular standard activation functions on different datasets and models on standard deep learning problems like image classification, object detection, semantic segmentation, and machine translation. The experimental results show that ErfAct and Pserf outperform in most networks compared to the standard activations. For all our experiments, we have first initialized the parameters $\alpha, \beta$ for ErfAct and $\gamma, \delta$ for Pserf and then updated via the backpropagation \cite{backp} algorithm (see \cite{prelu}) according to (\ref{eq7}) and for a single layer, the gradient of a parameter $\rho$ is:
\begin{align}\label{eq7}
    \frac{\partial L}{\partial \rho} = \sum_{x} \frac{\partial L}{\partial f(x)} \frac{\partial f(x)}{\partial \rho}
\end{align}
where L is the objective function, $\rho \in \{\alpha, \beta, \gamma, \delta\}$ and $f(x)\in \{\mathcal{F}_1(x;\alpha, \beta),  \mathcal{F}_2(x;\gamma ,\delta)$. 
For all of our experiments, to make a fair comparison between all the activations, we have first trained a network with hyper-parameter settings with the ReLU activation function and then only replaced ReLU with proposed activation functions and other baseline activations.

\subsection{Image Classification}
We present a detailed experimental comparison on MNIST \cite{mnist}, Fashion MNIST \cite{fashion}, SVHN \cite{SVHN}, CIFAR10 \cite{cifar10}, CIFAR100 \cite{cifar10}, Tiny ImageNet \cite{tiny}, and ImageNet-1k \cite{imagenet} dataset for image classification problem. We have trained the datasets with different standard models and report the top-1 accuracy. We have initialized the parameters $\alpha=0.75,\ \beta=0.75$ for ErfAct, and $\gamma=1.25,\ \delta=0.85$ for Pserf and update them according to (\ref{eq7}).
\subsubsection{MNIST, Fashion MNIST, and The Street View House Numbers (SVHN) Database:} We first evaluate our proposed activation functions on the MNIST \cite{mnist}, Fashion MNIST \cite{fashion}, and SVHN \cite{SVHN} datasets with AlexNet \cite{alexnet} and VGG-16 \cite{vgg} (with batch-normalization) models and results for 10-fold mean accuracy are reported in Table~\ref{tablll22} and Table~\ref{tabic} respectively. More detailed experiments on these datasets on LeNet \cite{lenet} and a custom-designed CNN architecture is reported in the supplementary material. We don't use any data augmentation for MNIST or Fashion MNIST, while we use standard data augmentation like rotation, zoom, height shift, shearing for the SVHN dataset. From Table~\ref{tablll22} and Table~\ref{tabic}, it is clear that the proposed functions outperformed all the baseline activation functions in all the three datasets and the performance are stable clear from mean$\pm$standard deviation.

\begin{table*}[!t]
\begin{center}
\small
\begin{tabular}{ |c|c|c|c|c| }
 \hline
 Activation Function &  \makecell{MNIST} & \makecell{Fashion MNIST} & \makecell{SVHN}  \\
 \hline
 ReLU  &  99.09 $\pm$ 0.10 & 93.22 $\pm$ 0.21 & 95.50 $\pm$ 0.22\\ 
 \hline
 
 Swish  & 99.30 $\pm$ 0.12  & 93.29 $\pm$ 0.22 & 95.59 $\pm$ 0.20\\
 \hline
 Leaky ReLU &  99.15 $\pm$ 0.13 & 93.30 $\pm$ 0.22 & 95.50 $\pm$ 0.28\\
 \hline
 ELU  & 99.29 $\pm$ 0.13  & 93.20 $\pm$ 0.25 & 95.60 $\pm$ 0.20\\
 \hline
 Softplus &  99.10 $\pm$ 0.14 & 93.18 $\pm$ 0.32 & 95.20 $\pm$ 0.37\\
 \hline
 Mish  &  99.27 $\pm$ 0.14 & 93.45 $\pm$ 0.32 & 95.60 $\pm$ 0.31\\
 \hline

 GELU & 99.22 $\pm$ 0.12 & 93.40 $\pm$ 0.25 & 95.55 $\pm$ 0.27\\
 \hline
 PAU & 99.31 $\pm$0.10 & 93.47 $\pm$ 0.23 & 95.67 $\pm$ 0.26 \\
 \hline
 PReLU & 99.15 $\pm$ 0.16  & 93.37 $\pm$ 0.31 & 95.42 $\pm$ 0.39\\
 \hline
 ReLU6 &  99.11 $\pm$ 0.10 & 93.26 $\pm$ 0.26 & 95.47 $\pm$ 0.24\\
 \hline
 ErfAct & \textbf{99.51} $\pm$ 0.10 & \textbf{93.79} $\pm$ 0.19 & \textbf{95.87} $\pm$ 0.20\\
 \hline
 Pserf& \textbf{99.49} $\pm$ 0.10& \textbf{93.82} $\pm$ 0.19 & \textbf{95.74} $\pm$ 0.22\\
 \hline
\end{tabular}
\caption{Comparison between different baseline activations and ErfAct and Pserf activations on MNIST, Fashion MNIST, and SVHN datasets in AlexNet. 10-fold mean accuracy (in \%) have been reported. mean$\pm$std is reported in the table.} 
\label{tablll22}
\end{center}
\end{table*}

\begin{table*}[!t]
\begin{center}
\small
\begin{tabular}{ |c|c|c|c|c| }
 \hline
 Activation Function &  \makecell{MNIST} & \makecell{Fashion MNIST} & \makecell{SVHN}  \\
 \hline
 ReLU  &  $99.05 \pm 0.11$ & $93.13 \pm 0.23$ & $95.09\pm 0.26$\\ 
 \hline
  Swish  & $99.09\pm 0.09$ & $93.34\pm 0.21$ & $95.29\pm 0.20$\\
 \hline
 Leaky ReLU & $99.02 \pm 0.14$ & $93.17\pm 0.28$ & $95.24\pm 0.23$\\
 \hline
 ELU  & $99.01\pm 0.15$ & $93.12\pm 0.30$ & $95.15\pm 0.28$\\
 \hline
 Softplus & $98.97\pm 0.14$ & $92.98\pm 0.34$ & $94.94\pm 0.30$\\
 \hline
Mish & 99.18 $\pm$ 0.07 & 93.47 $\pm$ 0.27 & 95.12 $\pm$ 0.25\\
 \hline
GELU & 99.10 $\pm$ 0.09 & 93.41 $\pm$ 0.29 & 95.11 $\pm$ 0.24\\
 \hline
 PAU & 99.07 $\pm$ 0.09 & 93.52 $\pm$ 0.24 & 95.23 $\pm$ 0.20 \\
 \hline
 PReLU & 99.01 $\pm$ 0.09 & 93.12 $\pm$ 0.27 & 95.14 $\pm$ 0.24\\
 \hline
 ReLU6 & 99.20 $\pm$ 0.08 & 93.25 $\pm$ 0.27 & 95.22 $\pm$ 0.20\\
 \hline
 ErfAct & \textbf{99.37} $\pm$ 0.06 & \textbf{93.81} $\pm$ 0.20 & \textbf{95.67} $\pm$ 0.18\\
 \hline
 Pserf& \textbf{99.38} $\pm$ 0.09 & \textbf{93.87} $\pm$ 0.22 & \textbf{95.66} $\pm$ 0.20\\
 \hline
\end{tabular}
\caption{Comparison between different baseline activations, ErfAct, and Pserf activations on MNIST, Fashion MNIST, and SVHN datasets on VGG-16 network. We report results for 10-fold mean accuracy (in \%). mean$\pm$std is reported in the table.} 
\label{tabic}
\end{center}
\end{table*}
\subsubsection{CIFAR:}
\begin{figure*}[!t]
    \begin{minipage}[t]{.495\linewidth}
        \centering
    
        \includegraphics[width=\linewidth]{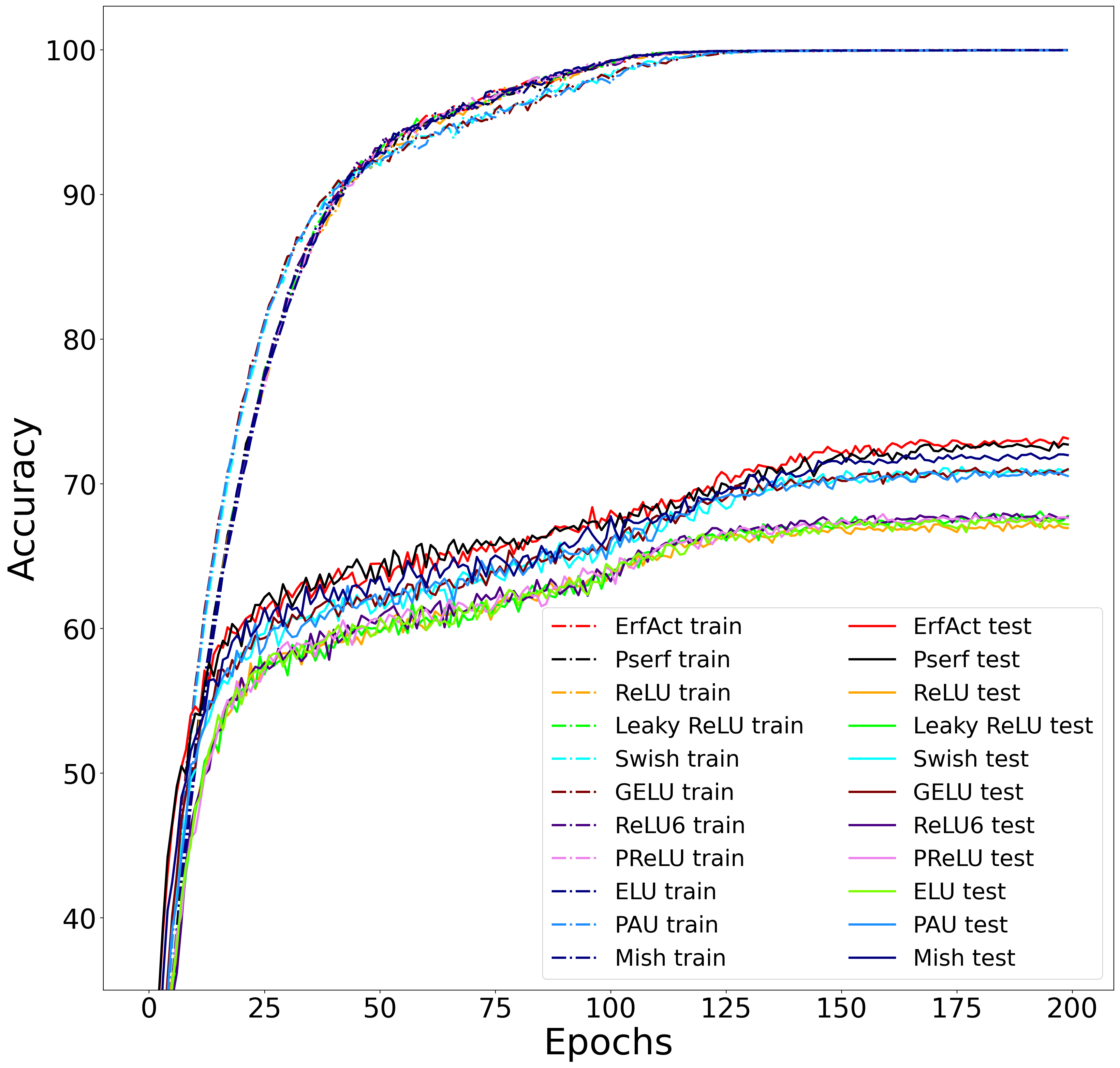}

        \caption{Top-1 Train and Test accuracy (higher is better) on CIFAR100 dataset with Shufflenet V2 (2.0x) network for different baseline activations, ErfAct, and Pserf.}
        \label{acc2}
    \end{minipage}
    \hfill
    \begin{minipage}[t]{.48\linewidth}
        \centering
        
       \includegraphics[width=\linewidth]{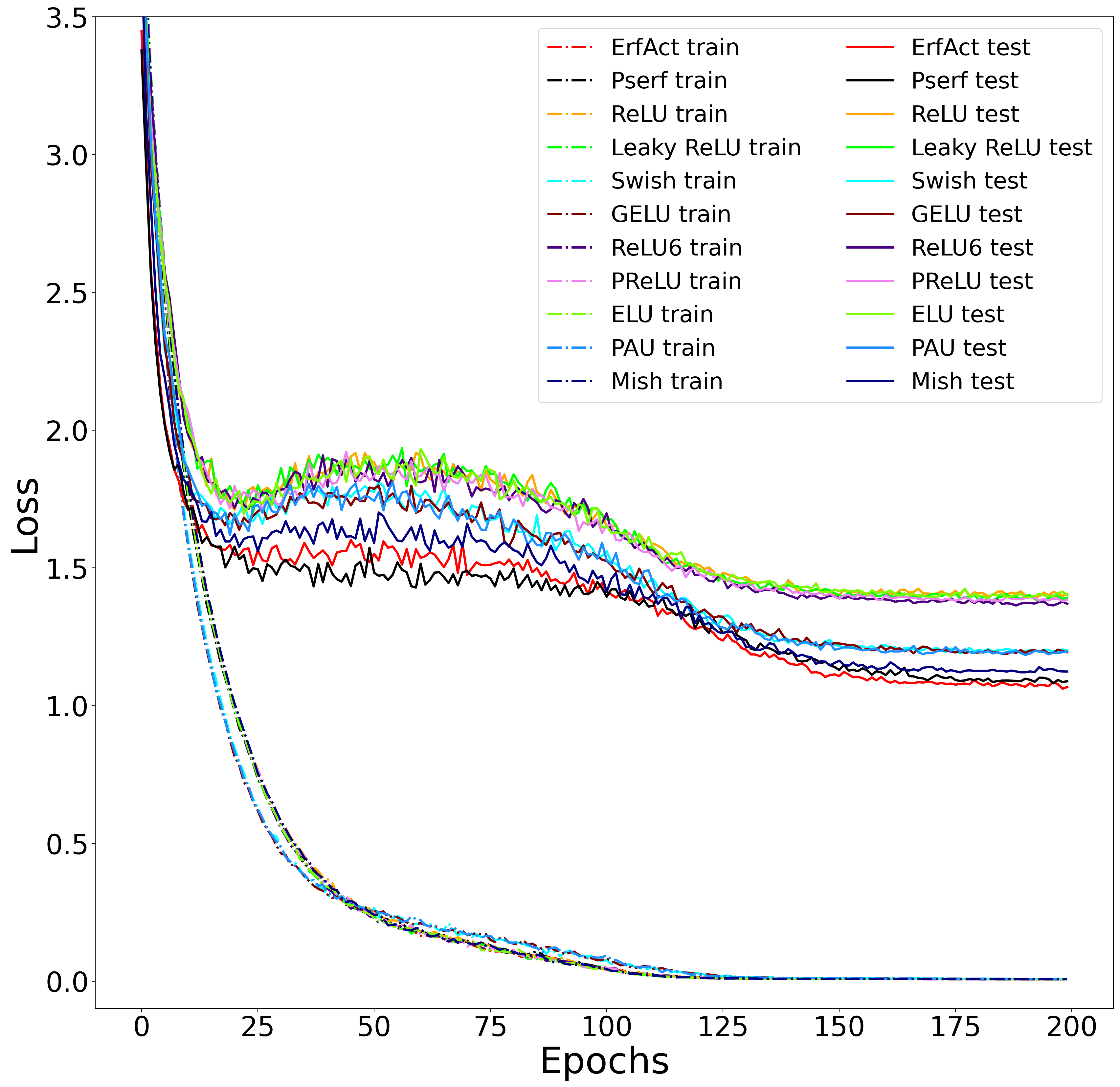}
      
        \caption{Top-1 Train and Test loss (lower is better) on CIFAR100 dataset with Shufflenet V2 (2.0x) network for different baseline activations, ErfAct, and Pserf.}
        \label{loss2}
    \end{minipage}  
\end{figure*}

Next we  have considered more challenging datasets like CIFAR100 and CIFAR10 to compare the performance of baseline activations and ErfAct and Pserf. We have reported the Top-1 accuracy for both the datasets for mean of 12 different runs on Table~\ref{tab3} and Table~\ref{tab2} with VGG-16 (with batch-normalization) \cite{vgg}, PreActResNet-34 (PA-ResNet-34) \cite{preactresnet},  Densenet-121 (DN-121) \cite{densenet}, MobileNet V2 (MN V2) \cite{mobile}, Resnet-50 \cite{resnet}, Inception V3 (IN-V3) \cite{incep}, WideResNet 28-10 (WRN 28-10) \cite{wrn}, and Shufflenet V2 (SF-V2 2.0x) \cite{shufflenet} models. A more detailed experiments on these datasets with EfficientNet B0 (EN-B0) \cite{efficientnet}, LeNet (LN) \cite{lenet}, AlexNet (AN) \cite{alexnet}, PreActResnet-18 (PARN-18) \cite{preactresnet}, Deep Layer Aggregation (DLA) \cite{dla}, Googlenet (GN) \cite{googlenet}, Resnext (Rxt) \cite{resnext}, Xception (Xpt) \cite{xception}, SimpleNet V1 (SN-V1) \cite{simple}, Squeezenet (SQ-Net) \cite{squeezenet}, ResNet18 (RN-18) \cite{resnet}, and Network in Network (NIN) \cite{nin} is reported in the supplementary section. From all the tables it is evident that the training is stable (mean$\pm$std) and the proposed activations archive 1\%-6\% higher top-1 accuracy in most of models compared to the baselines. The networks are trained upto 200 epochs with SGD optimizer \cite{sgd1, sgd2}, 0.9 momentum, and $5e^{-4}$ weight decay. We have started with 0.01 initial learning rate and decay the learning rate with cosine annealing \cite{cosan} learning rate scheduler. We consider batch size of 128. We consider standard data augmentation methods like horizontal flip, rotation for both the datasets. The Figures~\ref{acc2} and \ref{loss2} shows the learning curves on CIFAR100 dataset with Shufflenet V2 (2.0x) model for the baseline and the proposed activation functions and it is noticeable that training \& test accuracy curve is higher and loss curve is lower respectively for ErfAct and Pserf compared to the baseline activations.

\subsubsection{Tiny Imagenet:}
We consider a more challenging and important classification dataset Tiny Imagenet \cite{tiny} which is a similar type of dataset like ILSVRC and consisting of 200 classes with RGB images of size $64\times 64$ with total 1,00,000 training images, 10,000 validation images, and 10,000 test images. To compare the performance, we have considered WideResNet 28-10 (WRN 28-10) \cite{wrn} model and Top-1 accuracy is reported in table~\ref{tab22222} for mean of 5 different runs. 
\begin{table}[H]
\begin{center}
\begin{tabular}{ |c|c|c| }
 \hline
 Activation Function &  \makecell{Wide ResNet \\ 28-10 Model}  \\
 \hline
 ReLU  &  61.61 $\pm$ 0.47  \\ 
 \hline

 Swish  &  62.44 $\pm$ 0.49\\
  \hline
 Leaky ReLU &  61.47 $\pm$ 0.44 \\
 \hline
 ELU  &  61.99 $\pm$ 0.57\\
 \hline
 Softplus & 60.42 $\pm$ 0.61\\
  
 \hline
 Mish & 63.02 $\pm$ 0.57\\
 \hline
 GELU & 62.64 $\pm$ 0.62\\
 \hline
 PAU & 62.04 $\pm$ 0.54\\
 \hline
 PReLU & 61.25 $\pm$ 0.51\\
 \hline
 ReLU6 & 61.72 $\pm$ 0.56\\
  \hline
  
 ErfAct &  \textbf{64.20} $\pm$ 0.51\\ 
 \hline
 Pserf &  \textbf{64.01} $\pm$ 0.49 \\ 
  \hline
 \end{tabular}
\caption{Comparison between different baseline activations and ErfAct and Pserf on Tiny ImageNet dataset. Mean of 5 different runs for top-1 accuracy(in $\%$) have been reported. mean$\pm$std is reported in the table.} 
\label{tab22222}
\end{center}
\end{table}
\hspace*{-0.4cm}The model is trained with a batch size of 32, He Normal initializer \cite{prelu}, 0.2 dropout rate \cite{dropout}, adam optimizer \cite{adam}, with initial learning rate(lr rate) 0.01, and lr rate is reduced by a factor of 10 after every 60 epochs up-to 300 epochs. We have considered the standard data augmentation methods like rotation, width shift, height shift, shearing, zoom, horizontal flip, fill mode. From the table, it is clear that the performance for the proposed functions are better than the baseline functions and stable (mean$\pm$std) and got a boost in top-1 accuracy by $2.59\%$ and $2.40\%$ for ErfAct and Pserf compared to ReLU.

\begin{table*}[!htbp]
\begin{center}
\small
\begin{tabular}{ |c|c|c|c|c|c|c|c|c|c|c| }
 \hline
\makecell{Activation\\ Function} & VGG-16 & WRN 28-10 & ResNet-50 & PA-ResNet-34 & DN-121  & IN-V3 & MN-V2 & \makecell{SF-V2\\ 2.0x}\\

 \hline 
 ReLU & \makecell{71.67\\$\pm$0.28} & \makecell{76.32\\$\pm$0.25} &   \makecell{74.17\\$\pm$0.24} &  \makecell{73.12\\$\pm$0.23} & \makecell{75.67\\$\pm$0.28} & \makecell{74.23\\$\pm$0.26} & \makecell{74.02\\$\pm$0.24} & \makecell{67.49\\$\pm$0.26}\\ 
 \hline
 Leaky ReLU & \makecell{71.77\\$\pm$0.30} & \makecell{76.69\\$\pm$0.27} &   \makecell{74.11\\$\pm$0.27} &  \makecell{73.41\\$\pm$0.26} & \makecell{75.90\\$\pm$0.27} & \makecell{74.40\\$\pm$0.28} & \makecell{74.17\\$\pm$0.24} & \makecell{67.71\\$\pm$0.27}\\
 \hline
 ELU & \makecell{71.71\\$\pm$0.28} & \makecell{76.39\\$\pm$0.28} &   \makecell{74.51\\$\pm$0.24} &  \makecell{73.61\\$\pm$0.25} & \makecell{75.87\\$\pm$0.26} & \makecell{74.71\\$\pm$0.26} & \makecell{74.29\\$\pm$0.22} & \makecell{67.91\\$\pm$0.30}\\
 \hline
 Swish & \makecell{72.07\\$\pm$0.26} & \makecell{77.18\\$\pm$0.23} &   \makecell{75.10\\$\pm$0.24} &  \makecell{73.97\\$\pm$0.23} & \makecell{76.59\\$\pm$0.28} & \makecell{75.31\\$\pm$0.27} & \makecell{75.02\\$\pm$0.24} & \makecell{70.49\\$\pm$0.23}\\
 \hline
 Softplus & \makecell{71.10\\$\pm$0.32} & \makecell{75.36\\$\pm$0.37} &   \makecell{74.19\\$\pm$0.38} &  \makecell{73.17\\$\pm$0.36} & \makecell{75.08\\$\pm$0.36} & \makecell{74.20\\$\pm$0.34} & \makecell{74.33\\$\pm$0.38} & \makecell{68.93\\$\pm$0.36}\\
 \hline
 Mish & \makecell{72.31\\$\pm$0.24} & \makecell{77.40\\$\pm$0.25} &   \makecell{76.30\\$\pm$0.22} &  \makecell{75.14\\$\pm$0.21} & \makecell{77.11\\$\pm$0.25} & \makecell{76.22\\$\pm$0.25} & \makecell{75.31\\$\pm$0.21} & \makecell{71.79\\$\pm$0.22}\\
 \hline
 GELU  & \makecell{71.98\\$\pm$0.25} & \makecell{77.35\\$\pm$0.25} &   \makecell{75.61\\$\pm$0.22} &  \makecell{74.28\\$\pm$0.23} & \makecell{76.79\\$\pm$0.27} & \makecell{75.52\\$\pm$0.25} & \makecell{75.21\\$\pm$0.23} & \makecell{70.35\\$\pm$0.27}\\
 \hline
 PAU  & \makecell{71.72\\$\pm$0.25} & \makecell{77.20\\$\pm$0.26} &   \makecell{75.89\\$\pm$0.24} &  \makecell{74.41\\$\pm$0.23} & \makecell{76.59\\$\pm$0.28} & \makecell{75.79\\$\pm$0.28} & \makecell{75.07\\$\pm$0.19} & \makecell{70.68\\$\pm$0.26}\\
 \hline
 PReLU & \makecell{71.77\\$\pm$0.30} & \makecell{76.79\\$\pm$0.27} &   \makecell{74.45\\$\pm$0.29} &  \makecell{73.32\\$\pm$0.27} & \makecell{76.19\\$\pm$0.30} & \makecell{74.51\\$\pm$0.29} & \makecell{74.31\\$\pm$0.32} & \makecell{68.35\\$\pm$0.30}\\
 \hline
 ReLU6 & \makecell{72.07\\$\pm$0.27} & \makecell{76.62\\$\pm$0.28} &   \makecell{74.37\\$\pm$0.24} &  \makecell{73.50\\$\pm$0.24} & \makecell{76.07\\$\pm$0.26} & \makecell{74.69\\$\pm$0.25} & \makecell{74.64\\$\pm$0.24} & \makecell{67.93\\$\pm$0.26}\\
 \hline
 ErfAct & \makecell{\textbf{72.93}\\$\pm$0.22} & \makecell{\textbf{78.49}\\$\pm$0.23} &   \makecell{\textbf{77.09}\\$\pm$0.20} &  \makecell{\textbf{76.21}\\$\pm$0.20} & \makecell{\textbf{78.18}\\$\pm$0.23} & \makecell{\textbf{77.12}\\$\pm$0.24} & \makecell{\textbf{76.23}\\$\pm$0.19} & \makecell{\textbf{73.17}\\$\pm$0.22}\\ 
 \hline
 Pserf & \makecell{\textbf{72.69}\\$\pm$0.24} & \makecell{\textbf{78.31}\\$\pm$0.24} &   \makecell{\textbf{76.97}\\$\pm$0.20} &  \makecell{\textbf{75.91}\\$\pm$0.22} & \makecell{\textbf{78.38}\\$\pm$0.22} & \makecell{\textbf{77.01}\\$\pm$0.25} & \makecell{\textbf{76.07}\\$\pm$0.21} & \makecell{\textbf{72.91}\\$\pm$0.21}\\ 
 \hline
  
 \end{tabular}
\caption{Comparison between different baseline activations and ErfAct and Pserf on CIFAR100 dataset. Top-1 accuracy(in $\%$) for mean of 12 different runs have been reported. mean$\pm$std is reported in the table.} 
\label{tab3}
\end{center}
\end{table*}

\begin{table*}[!htbp]
\begin{center}
\small
\begin{tabular}{ |c|c|c|c|c|c|c|c|c|c|c| }
 \hline
\makecell{Activation\\ Function} & VGG-16 & WRN 28-10 & ResNet-50 & PA-ResNet-34 & DN-121  & IN-V3 & MN-V2 & \makecell{SF-V2\\ 2.0x}\\

 \hline 
 ReLU & \makecell{93.44\\$\pm$0.22} & \makecell{95.17\\$\pm$0.21} &   \makecell{94.35\\$\pm$0.18} &  \makecell{94.17\\$\pm$0.19} & \makecell{94.77\\$\pm$0.20} & \makecell{94.15\\$\pm$0.20} & \makecell{94.20\\$\pm$0.16} & \makecell{91.63\\$\pm$0.21}\\ 
 \hline
 Leaky ReLU & \makecell{93.65\\$\pm$0.21} & \makecell{95.02\\$\pm$0.22} &   \makecell{94.45\\$\pm$0.20} &  \makecell{94.33\\$\pm$0.18} & \makecell{94.89\\$\pm$0.22} & \makecell{94.20\\$\pm$0.22} & \makecell{94.32\\$\pm$0.19} & \makecell{91.82\\$\pm$0.23}\\
 \hline
 ELU & \makecell{93.70\\$\pm$0.19} & \makecell{95.28\\$\pm$0.20} &   \makecell{94.27\\$\pm$0.24} &  \makecell{94.30\\$\pm$0.25} & \makecell{94.64\\$\pm$0.18} & \makecell{94.38\\$\pm$0.17} & \makecell{94.27\\$\pm$0.18} & \makecell{91.99\\$\pm$0.20}\\
 \hline
 Swish & \makecell{93.77\\$\pm$0.18} & \makecell{95.41\\$\pm$0.17} &  \makecell{94.61\\$\pm$0.24} &  \makecell{94.47\\$\pm$0.25} & \makecell{94.81\\$\pm$0.19} & \makecell{94.51\\$\pm$0.17} & \makecell{94.40\\$\pm$0.20} & \makecell{92.17\\$\pm$0.25}\\
 \hline
 Softplus & \makecell{93.10\\$\pm$0.33} & \makecell{94.77\\$\pm$0.30} &  \makecell{93.91\\$\pm$0.30} &  \makecell{94.07\\$\pm$0.35} & \makecell{94.41\\$\pm$0.34} & \makecell{94.21\\$\pm$0.32} & \makecell{93.79\\$\pm$0.29} & \makecell{91.32\\$\pm$0.33}\\
 \hline
 Mish & \makecell{93.91\\$\pm$0.17} & \makecell{95.35\\$\pm$0.18} &  \makecell{94.78\\$\pm$0.22} &  \makecell{94.55\\$\pm$0.23} & \makecell{95.03\\$\pm$0.15} & \makecell{94.64\\$\pm$0.18} & \makecell{94.71\\$\pm$0.18} & \makecell{92.41\\$\pm$0.20}\\
 \hline
 GELU & \makecell{93.71\\$\pm$0.17} & \makecell{95.28\\$\pm$0.19} &  \makecell{94.64\\$\pm$0.23} &  \makecell{94.31\\$\pm$0.25} & \makecell{94.99\\$\pm$0.19} & \makecell{94.57\\$\pm$0.21} & \makecell{94.40\\$\pm$0.18} & \makecell{92.27\\$\pm$0.20}\\
 \hline
PAU & \makecell{93.57\\$\pm$0.22} & \makecell{95.27\\$\pm$0.20} &  \makecell{94.67\\$\pm$0.23} &  \makecell{94.41\\$\pm$0.24} & \makecell{94.74\\$\pm$0.20} & \makecell{94.57\\$\pm$0.19} & \makecell{94.51\\$\pm$0.14} & \makecell{92.30\\$\pm$0.21}\\
 \hline
 PReLU & \makecell{93.41\\$\pm$0.23} & \makecell{95.02\\$\pm$0.24} &   \makecell{94.27\\$\pm$0.26} &  \makecell{94.30\\$\pm$0.26} &
 \makecell{94.51\\$\pm$0.24} & 
 \makecell{94.49\\$\pm$0.22} & 
 \makecell{94.32\\$\pm$0.23} & \makecell{91.80\\$\pm$0.25}\\ 
 \hline
 ReLU6 & \makecell{93.72\\$\pm$0.17} & \makecell{95.32\\$\pm$0.19} &   \makecell{94.30\\$\pm$0.24} &  \makecell{94.21\\$\pm$0.24} & 
 \makecell{94.61\\$\pm$0.20} & 
 \makecell{94.42\\$\pm$0.20} & 
 \makecell{94.18\\$\pm$0.19} & \makecell{91.71\\$\pm$0.21}\\ 
 \hline
 ErfAct & \makecell{\textbf{94.47}\\$\pm$0.15} & \makecell{\textbf{95.88}\\$\pm$0.12} &  \makecell{\textbf{95.01}\\$\pm$0.17} &  \makecell{\textbf{95.21}\\$\pm$0.18} & \makecell{\textbf{95.71}\\$\pm$0.15}  & \makecell{\textbf{95.29}\\$\pm$0.14} & \makecell{\textbf{95.34}\\$\pm$0.12} & \makecell{\textbf{93.74}\\$\pm$0.18}\\ 
 \hline
 Pserf & \makecell{\textbf{94.24}\\$\pm$0.16} & \makecell{\textbf{95.71}\\$\pm$0.13} &  \makecell{\textbf{95.14}\\$\pm$0.19} &  \makecell{\textbf{95.08}\\$\pm$0.29} & \makecell{\textbf{95.62}\\$\pm$0.17}  & \makecell{\textbf{95.10}\\$\pm$0.13} & \makecell{\textbf{95.19}\\$\pm$0.14} & \makecell{\textbf{93.59}\\$\pm$0.18}\\
 \hline
  
 \end{tabular}
\caption{Comparison between different baseline activations and ErfAct and Pserf on CIFAR10 dataset. Top-1 accuracy(in $\%$) for mean of 12 different runs have been reported. mean$\pm$std is reported in the table.} 
\label{tab2}
\end{center}
\end{table*}


\subsection{\textbf{Semantic Segmentation}}
Semantic segmentation is an important problem in deep learning. In this section, we present experimental results on the Cityscapes dataset \cite{city}. We report the pixel accuracy and mean Intersection-Over-Union (mIOU) on the U-net model\cite{unet}.  The model is trained up to 250 epochs, with adam optimizer \cite{adam}, learning rate 5$e^{-3}$, batch size 32 and Xavier Uniform initializer \cite{xavier}. A mean of 5 different runs on the test dataset is reported in table~\ref{tabsg}. We got around $1.97\%$ and $1.89\%$ boost on mIOU for ErfAct and Pserf compared to ReLU.

\begin{table}[H]
\begin{center}
\begin{tabular}{ |c|c|c|c| }
 \hline
 Activation Function & \makecell{
Pixel\\ Accuracy} & mIOU  \\
 \hline
 ReLU  & 79.60 $\pm$ 0.45 & 69.32 $\pm$ 0.30 \\ 
 \hline
 Swish  & 79.71 $\pm$ 0.49 & 69.68 $\pm$ 0.31\\
 \hline
 Leaky ReLU & 79.41 $\pm$ 0.42 & 69.48 $\pm$ 0.39\\
 \hline
 ELU  & 79.27 $\pm$ 0.54 & 68.12 $\pm$ 0.41\\
 \hline
 Softplus & 78.69 $\pm$ 0.49 & 68.12 $\pm$ 0.55\\
 \hline
 Mish & 80.12 $\pm$ 0.45 & 69.87 $\pm$ 0.29\\
 \hline
 GELU & 79.60 $\pm $0.39 & 69.51 $\pm$ 0.39\\
 \hline
 PAU & 79.95 $\pm $ 0.41 & 69.42 $\pm $ 0.46\\
 \hline
 PReLU & 78.99 $\pm$ 0.42 & 68.82 $\pm$ 0.41\\
 \hline
 ReLU6 & 79.59 $\pm$ 0.41 & 69.66 $\pm$ 0.41\\
 \hline
ErfAct & \textbf{81.41} $\pm$ 0.45 & \textbf{71.29} $\pm$ 0.31 \\ 
\hline
Pserf & \textbf{81.12} $\pm$ 0.42 & \textbf{71.21} $\pm$ 0.34 \\ 
 \hline
 \end{tabular}
\caption{Comparison between different baseline activations and ErfAct and Pserf on semantic segmentation problem on U-NET model in CityScapes dataset. mean$\pm$std is reported in the table.} 
\label{tabsg}
\end{center}
\end{table}
\subsection{\textbf{Object Detection}}
Object detection is a standard problem in computer vision. In this section, we have reported our experimental results on challenging Pascal VOC dataset \cite{pascal} with Single Shot MultiBox Detector(SSD) 300 \cite{ssd} with VGG-16(with batch-normalization) \cite{vgg} as the backbone network. The mean average precision (mAP) is reported in Table~\ref{tabod} for a mean of 8 different runs. The model is trained with batch size of 8, 0.001 learning rate, SGD optimizer \cite{sgd1, sgd2} with 0.9 momentum, 5$e^{-4}$ weight decay for 120000 iterations. The results are stable on different runs (mean$\pm$std). We got around $1\%$ boost in mAP for both ErfAct and Pserf compared to ReLU.

\begin{table}[H]
\begin{center}
\begin{tabular}{ |c|c|c| }
 \hline
 Activation Function &  \makecell{mAP}  \\
  
 \hline
 ReLU  &  77.2 $\pm$ 0.14  \\ 
 \hline
 Swish  &  77.5 $\pm$ 0.12\\
 \hline
 Leaky ReLU &  77.2 $\pm$ 0.19 \\
 \hline
 ELU  &  75.1 $\pm$ 0.22\\
 \hline
 Softplus & 74.2 $\pm$ 0.25\\
 \hline
 Mish & 77.6 $\pm$ 0.14\\
 \hline
 GELU & 77.5 $\pm$ 0.14\\
 \hline
 PAU & 77.4 $\pm$ 0.16\\
 \hline
PReLU & 77.2 $\pm$ 0.20\\
\hline
ReLU6 & 77.1 $\pm$ 0.15\\
\hline
ErfAct &  \textbf{78.2} $\pm$ 0.12 \\ 
 \hline
Pserf &  \textbf{78.2} $\pm$ 0.14 \\
\hline
 \end{tabular}
\caption{Comparison between different baseline activations and ErfAct and Pserf on Object Detection problem on SSD 300 model in Pascal-VOC dataset. mean$\pm$std is reported in the table.} 
\label{tabod}
\end{center}
\end{table}

.

\subsection{\textbf{Machine Translation}}

\begin{table}[H]
\begin{center}
\begin{tabular}{ |c|c|c| }
 \hline
 Activation Function & \makecell{
BLEU Score on\\ the newstest2014 dataset }  \\
 
 \hline
 ReLU  &  26.2 $\pm$ 0.15  \\ 
 \hline
 Swish  &  26.4 $\pm$ 0.10\\
 \hline
 Leaky ReLU &  26.3 $\pm$ 0.17 \\
 \hline
 ELU  &  25.1 $\pm$ 0.15\\
 \hline
 Softplus & 23.6 $\pm$ 0.16\\
 \hline
 Mish & 26.3 $\pm$ 0.12\\
 \hline
 GELU & 26.4 $\pm$ 0.19\\
 \hline
 PAU & 26.3 $\pm$ 0.16\\
 \hline
  PReLU & 26.2 $\pm$ 0.21\\
 \hline
 ReLU6 & 26.1 $\pm$ 0.14\\
 \hline
 ErfAct &  \textbf{26.8} $\pm$ 0.11 \\ 
 \hline
 Pserf &  \textbf{26.7} $\pm$ 0.10 \\ 
 \hline
 \end{tabular}
\caption{Comparison between different baseline activations and ErfAct and Pserf on Machine translation problem on transformer model in WMT-2014 dataset. mean$\pm$std is reported in the table.} 
\label{tabmt}
\end{center}
\end{table}
Machine Translation is a procedure in which text or speech is translated from one language to another language without the help of any human being. We consider the standard WMT 2014 English$\rightarrow$German dataset for our experiment. The database contains 4.5 million training sentences. We train an attention-based 8-head transformer network \cite{attn} with Adam optimizer \cite{adam}, 0.1 dropout rate \cite{dropout}, and train up to 100000 steps. We try to keep other hyperparameters similar as mentioned in the original paper \cite{attn}. We evaluate the network performance on the newstest2014 dataset using the BLEU score metric. The mean of 5 different runs is being reported on Table~\ref{tabmt} on the test dataset(newstest2014). From the table, it is clear that the results are stable on different runs (mean$\pm$std), and we got around $0.6\%$ and $0.5\%$ boost in BLEU score for ErfAct and Pserf compared to ReLU. 
\section{Baseline Table}

\begin{table*}[!t]
\newenvironment{amazingtabular}{\begin{tabular}{*{50}{l}}}{\end{tabular}}
\centering
\begin{amazingtabular}
\midrule
Baselines & ReLU & \makecell{Leaky\\ ReLU} & ELU & Softplus & Swish  & PReLU & ReLU6 & Mish & GELU & PAU\\
\midrule
ErfAct $>$ \text{Baseline} & \hspace{0.3cm}57 & \hspace{0.3cm}57 & \hspace{0.3cm}57 & \hspace{0.3cm}57 & \hspace{0.3cm}54 & \hspace{0.3cm}56 & \hspace{0.3cm}56 & \hspace{0.3cm}53 & \hspace{0.3cm}55 & \hspace{0.3cm}55\\
ErfAct $=$ \text{Baseline} & \hspace{0.3cm}0 & \hspace{0.3cm}0 & \hspace{0.3cm}0 & \hspace{0.3cm}0 & \hspace{0.3cm}0 & \hspace{0.3cm}0  & \hspace{0.3cm}0  & \hspace{0.3cm}0  & \hspace{0.3cm}0 & \hspace{0.3cm}0\\
ErfAct $<$ \text{Baseline} & \hspace{0.3cm}0 & \hspace{0.3cm}0 & \hspace{0.3cm}0 & \hspace{0.3cm}0 & \hspace{0.3cm}3 & \hspace{0.3cm}1 & \hspace{0.3cm}1 & \hspace{0.3cm}4 & \hspace{0.3cm}2 & \hspace{0.3cm}2 \\
\midrule
Pserf $>$ \text{Baseline} & \hspace{0.3cm}57 & \hspace{0.3cm}57 & \hspace{0.3cm}57 & \hspace{0.3cm}57 & \hspace{0.3cm}54 & \hspace{0.3cm}56 & \hspace{0.3cm}56 & \hspace{0.3cm}53 & \hspace{0.3cm}55 & \hspace{0.3cm}54\\
Pserf $=$ \text{Baseline} & \hspace{0.3cm}0 & \hspace{0.3cm}0 & \hspace{0.3cm}0 & \hspace{0.3cm}0 & \hspace{0.3cm}0 & \hspace{0.3cm}0  & \hspace{0.3cm}0  & \hspace{0.3cm}0  & \hspace{0.3cm}0 & \hspace{0.3cm}0\\
Pserf $<$ \text{Baseline} & \hspace{0.3cm}0 & \hspace{0.3cm}0 & \hspace{0.3cm}0 & \hspace{0.3cm}0 & \hspace{0.3cm}3 & \hspace{0.3cm}1 & \hspace{0.3cm}1 & \hspace{0.3cm}4 & \hspace{0.3cm}2 & \hspace{0.3cm}3 \\
\midrule
\end{amazingtabular}
  \caption{Baseline table for ErfAct and Pserf. These numbers represent the total number of models in which ErfAct and Pserf underperforms, equal or outperforms compared to the baseline activation functions}
  \label{tab49}
\end{table*}
The experiment section shows that ErfAct and Pserf beat or perform equally well with baseline activation functions in most cases while under-performs marginally on rare occasions. We provide a detailed comparison based on all the experiments in earlier sections and supplementary material with the proposed and the baseline activation functions in Table~\ref{tab49}.
\subsection{\textbf{Computational Time Comparison}}
We present the time comparison for the baseline activation functions and ErfAct, Pserf for the mean of 100 runs for both forward and backward pass on a $32\times 32$ RGB image in PreActResNet-18 \cite{preactresnet} model in Table~\ref{tabtime}. An NVIDIA Tesla V100 GPU with 32GB ram is used to run the experiments.
From Table~\ref{tabtime} and the experiment section, it is clear that there is a small trade-off between the computational time and the model performance when compared to ReLU as the proposed activations contain trainable parameters. In contrast, the time is comparable with Swish, Mish or GELU \& much better than PAU and model performance

\begin{table}[H]
\begin{center}
\begin{tabular}{ |c|c|c|c| }
 \hline
 \makecell{Activation\\ Function} &  Forward Pass & Backward Pass \\
 
 \hline
 ReLU  &  
5.39 $\pm$ 0.39 $\mu$s &
5.70 $\pm$ 1.56 $\mu$s
  \\ 
 \hline
 Swish  &  
8.35 $\pm$ 1.44 $\mu$s &
10.56 $\pm$ 2.34 $\mu$s
\\
 \hline
Leaky ReLU &  
5.50 $\pm$ 0.51 $\mu$s &
5.97 $\pm$ 0.75 $\mu$s
 \\
 \hline
 ELU  &  
6.17 $\pm$ 0.50 $\mu$s &
5.93 $\pm$ 0.93 $\mu$s
\\
 \hline
 Softplus & 
6.13 $\pm$ 0.49 $\mu$s &
5.94 $\pm$ 0.55 $\mu$s
\\
 \hline
 Mish & 
7.45 $\pm$ 2.55 $\mu$s &
8.89 $\pm$ 2.85 $\mu$s
\\
 \hline
 GELU & 
8.87 $\pm$ 1.54 $\mu$s &
9.22 $\pm$ 1.75 $\mu$s
\\
 \hline
 PAU & 19.05 $\pm$ 2.69 $\mu$s & 32.62 $\pm$ 3.76 $\mu$s\\
 \hline
 PReLU & 
6.12 $\pm$ 0.90 $\mu$s &
6.23 $\pm$ 1.41 $\mu$s
\\
 \hline
 ReLU6 & 
5.77$\pm$ 0.73 $\mu$s &
5.73$\pm$ 0.66 $\mu$s
\\
 \hline
 
 ErfAct &  
7.41 $\pm$ 1.51 $\mu$s 
& 10.62 $\pm$ 1.53 $\mu$s
 \\ 
 \hline
 Pserf &  
7.53 $\pm$ 1.77 $\mu$s
& 10.77 $\pm$ 1.78 $\mu$s\\ 
\hline
 \end{tabular}
\caption{Runtime comparison for the forward and backward passes for ErfAct and Pserf and baseline activation functions for a 32$\times$ 32 RGB image in PreActResNet-18 model.} 
\label{tabtime}
\end{center}
\end{table}
\hspace*{-0.35cm}comparatively much better than baseline activations in most cases.

\section{Conclusion}
In this work, we propose two simple and effective novel activation functions. We call them ErfAct and Pserf. The method of construction is a combination of functions using trainable parameters. The proposed functions are unbounded above, bounded below, non-monotonic, smooth and zero centred. We show that both functions can approximate the ReLU activation function. Across most of the experiments, ErfAct and Pserf are top-performing activation functions from which we can conclude that the proposed functions have the potential to replace the widely used activations like ReLU, Swish or Mish.
\section{Acknowledgements}
The authors would like to express their gratitude to Dr. Bapi Chatterjee for lending GPU equipment which helped in carrying out some of the important experiments. 

\bibliography{aaai22}

\begin{thebibliography}{57}
\providecommand{\natexlab}[1]{#1}

\bibitem[{Biswas, Banerjee, and Pandey(2021)}]{opau}
Biswas, K.; Banerjee, S.; and Pandey, A.~K. 2021.
\newblock Orthogonal-Pad\'e Activation Functions: Trainable Activation
  functions for smooth and faster convergence in deep networks.
\newblock arXiv:2106.09693.

\bibitem[{Bochkovskiy, Wang, and Liao(2020)}]{yolo}
Bochkovskiy, A.; Wang, C.-Y.; and Liao, H.-Y.~M. 2020.
\newblock YOLOv4: Optimal Speed and Accuracy of Object Detection.
\newblock arXiv:2004.10934.

\bibitem[{Brown et~al.(2020)Brown, Mann, Ryder, Subbiah, Kaplan, Dhariwal,
  Neelakantan, Shyam, Sastry, Askell, Agarwal, Herbert-Voss, Krueger, Henighan,
  Child, Ramesh, Ziegler, Wu, Winter, Hesse, Chen, Sigler, Litwin, Gray, Chess,
  Clark, Berner, McCandlish, Radford, Sutskever, and Amodei}]{gpt3}
Brown, T.~B.; Mann, B.; Ryder, N.; Subbiah, M.; Kaplan, J.; Dhariwal, P.;
  Neelakantan, A.; Shyam, P.; Sastry, G.; Askell, A.; Agarwal, S.;
  Herbert-Voss, A.; Krueger, G.; Henighan, T.; Child, R.; Ramesh, A.; Ziegler,
  D.~M.; Wu, J.; Winter, C.; Hesse, C.; Chen, M.; Sigler, E.; Litwin, M.; Gray,
  S.; Chess, B.; Clark, J.; Berner, C.; McCandlish, S.; Radford, A.; Sutskever,
  I.; and Amodei, D. 2020.
\newblock Language Models are Few-Shot Learners.
\newblock arXiv:2005.14165.

\bibitem[{Chollet(2017)}]{xception}
Chollet, F. 2017.
\newblock Xception: Deep Learning with Depthwise Separable Convolutions.
\newblock arXiv:1610.02357.

\bibitem[{Clevert, Unterthiner, and Hochreiter(2016)}]{elu}
Clevert, D.-A.; Unterthiner, T.; and Hochreiter, S. 2016.
\newblock Fast and Accurate Deep Network Learning by Exponential Linear Units
  (ELUs).
\newblock arXiv:1511.07289.

\bibitem[{Cordts et~al.(2016)Cordts, Omran, Ramos, Rehfeld, Enzweiler,
  Benenson, Franke, Roth, and Schiele}]{city}
Cordts, M.; Omran, M.; Ramos, S.; Rehfeld, T.; Enzweiler, M.; Benenson, R.;
  Franke, U.; Roth, S.; and Schiele, B. 2016.
\newblock The Cityscapes Dataset for Semantic Urban Scene Understanding.
\newblock arXiv:1604.01685.

\bibitem[{Deng et~al.(2009)Deng, Dong, Socher, Li, Li, and Fei-Fei}]{imagenet}
Deng, J.; Dong, W.; Socher, R.; Li, L.-J.; Li, K.; and Fei-Fei, L. 2009.
\newblock ImageNet: A large-scale hierarchical image database.
\newblock In \emph{2009 IEEE Conference on Computer Vision and Pattern
  Recognition}, 248--255.

\bibitem[{Elfwing, Uchibe, and Doya(2017)}]{silu}
Elfwing, S.; Uchibe, E.; and Doya, K. 2017.
\newblock Sigmoid-Weighted Linear Units for Neural Network Function
  Approximation in Reinforcement Learning.
\newblock arXiv:1702.03118.

\bibitem[{Everingham et~al.(2010)Everingham, Gool, Williams, Winn, and
  Zisserman}]{pascal}
Everingham, M.; Gool, L.; Williams, C.~K.; Winn, J.; and Zisserman, A. 2010.
\newblock The Pascal Visual Object Classes (VOC) Challenge.
\newblock \emph{Int. J. Comput. Vision}, 88(2): 303–338.

\bibitem[{Glorot and Bengio(2010)}]{xavier}
Glorot, X.; and Bengio, Y. 2010.
\newblock Understanding the difficulty of training deep feedforward neural
  networks.
\newblock In Teh, Y.~W.; and Titterington, M., eds., \emph{Proceedings of the
  Thirteenth International Conference on Artificial Intelligence and
  Statistics}, volume~9 of \emph{Proceedings of Machine Learning Research},
  249--256. Chia Laguna Resort, Sardinia, Italy: JMLR Workshop and Conference
  Proceedings.

\bibitem[{HasanPour et~al.(2016)HasanPour, Rouhani, Fayyaz, and
  Sabokrou}]{simple}
HasanPour, S.~H.; Rouhani, M.; Fayyaz, M.; and Sabokrou, M. 2016.
\newblock Lets keep it simple, Using simple architectures to outperform deeper
  and more complex architectures.
\newblock \emph{CoRR}, abs/1608.06037.

\bibitem[{He et~al.(2015{\natexlab{a}})He, Zhang, Ren, and Sun}]{resnet}
He, K.; Zhang, X.; Ren, S.; and Sun, J. 2015{\natexlab{a}}.
\newblock Deep Residual Learning for Image Recognition.
\newblock arXiv:1512.03385.

\bibitem[{He et~al.(2015{\natexlab{b}})He, Zhang, Ren, and Sun}]{prelu}
He, K.; Zhang, X.; Ren, S.; and Sun, J. 2015{\natexlab{b}}.
\newblock Delving Deep into Rectifiers: Surpassing Human-Level Performance on
  ImageNet Classification.
\newblock arXiv:1502.01852.

\bibitem[{He et~al.(2016)He, Zhang, Ren, and Sun}]{preactresnet}
He, K.; Zhang, X.; Ren, S.; and Sun, J. 2016.
\newblock Identity Mappings in Deep Residual Networks.
\newblock arXiv:1603.05027.

\bibitem[{Hendrycks and Gimpel(2020)}]{gelu}
Hendrycks, D.; and Gimpel, K. 2020.
\newblock Gaussian Error Linear Units (GELUs).
\newblock arXiv:1606.08415.

\bibitem[{Huang et~al.(2016)Huang, Liu, van~der Maaten, and
  Weinberger}]{densenet}
Huang, G.; Liu, Z.; van~der Maaten, L.; and Weinberger, K.~Q. 2016.
\newblock Densely Connected Convolutional Networks.
\newblock arXiv:1608.06993.

\bibitem[{Iandola et~al.(2016)Iandola, Han, Moskewicz, Ashraf, Dally, and
  Keutzer}]{squeezenet}
Iandola, F.~N.; Han, S.; Moskewicz, M.~W.; Ashraf, K.; Dally, W.~J.; and
  Keutzer, K. 2016.
\newblock SqueezeNet: AlexNet-level accuracy with 50x fewer parameters and
  <0.5MB model size.
\newblock arXiv:1602.07360.

\bibitem[{Ioffe and Szegedy(2015)}]{batch}
Ioffe, S.; and Szegedy, C. 2015.
\newblock Batch Normalization: Accelerating Deep Network Training by Reducing
  Internal Covariate Shift.
\newblock arXiv:1502.03167.

\bibitem[{Kidger and Lyons(2020)}]{universal}
Kidger, P.; and Lyons, T. 2020.
\newblock Universal Approximation with Deep Narrow Networks.
\newblock arXiv:1905.08539.

\bibitem[{Kiefer and Wolfowitz(1952)}]{sgd2}
Kiefer, J.; and Wolfowitz, J. 1952.
\newblock Stochastic Estimation of the Maximum of a Regression Function.
\newblock \emph{Annals of Mathematical Statistics}, 23: 462--466.

\bibitem[{Kingma and Ba(2015)}]{adam}
Kingma, D.~P.; and Ba, J. 2015.
\newblock Adam: {A} Method for Stochastic Optimization.
\newblock In Bengio, Y.; and LeCun, Y., eds., \emph{3rd International
  Conference on Learning Representations, {ICLR} 2015, San Diego, CA, USA, May
  7-9, 2015, Conference Track Proceedings}.

\bibitem[{Krizhevsky(2009)}]{cifar10}
Krizhevsky, A. 2009.
\newblock Learning multiple layers of features from tiny images.
\newblock Technical report, University of Toronto.

\bibitem[{Krizhevsky(2010)}]{relu6}
Krizhevsky, A. 2010.
\newblock Convolutional deep belief networks on cifar-10.

\bibitem[{Krizhevsky, Sutskever, and Hinton(2012)}]{alexnet}
Krizhevsky, A.; Sutskever, I.; and Hinton, G.~E. 2012.
\newblock ImageNet Classification with Deep Convolutional Neural Networks.
\newblock In \emph{Proceedings of the 25th International Conference on Neural
  Information Processing Systems - Volume 1}, NIPS'12, 1097–1105. Red Hook,
  NY, USA: Curran Associates Inc.

\bibitem[{Le and Yang(2015)}]{tiny}
Le, Y.; and Yang, X. 2015.
\newblock Tiny ImageNet Visual Recognition Challenge.

\bibitem[{{LeCun} et~al.(1989){LeCun}, {Boser}, {Denker}, {Henderson},
  {Howard}, {Hubbard}, and {Jackel}}]{backp}
{LeCun}, Y.; {Boser}, B.; {Denker}, J.~S.; {Henderson}, D.; {Howard}, R.~E.;
  {Hubbard}, W.; and {Jackel}, L.~D. 1989.
\newblock Backpropagation Applied to Handwritten Zip Code Recognition.
\newblock \emph{Neural Computation}, 1(4): 541--551.

\bibitem[{Lecun et~al.(1998)Lecun, Bottou, Bengio, and Haffner}]{lenet}
Lecun, Y.; Bottou, L.; Bengio, Y.; and Haffner, P. 1998.
\newblock Gradient-based learning applied to document recognition.
\newblock \emph{Proceedings of the IEEE}, 86(11): 2278--2324.

\bibitem[{LeCun, Cortes, and Burges(2010)}]{mnist}
LeCun, Y.; Cortes, C.; and Burges, C. 2010.
\newblock MNIST handwritten digit database.
\newblock \emph{ATT Labs [Online]. Available:
  http://yann.lecun.com/exdb/mnist}, 2.

\bibitem[{Lin, Chen, and Yan(2014)}]{nin}
Lin, M.; Chen, Q.; and Yan, S. 2014.
\newblock Network In Network.
\newblock arXiv:1312.4400.

\bibitem[{Lin et~al.(2015)Lin, Maire, Belongie, Bourdev, Girshick, Hays,
  Perona, Ramanan, Zitnick, and Dollár}]{coco}
Lin, T.-Y.; Maire, M.; Belongie, S.; Bourdev, L.; Girshick, R.; Hays, J.;
  Perona, P.; Ramanan, D.; Zitnick, C.~L.; and Dollár, P. 2015.
\newblock Microsoft COCO: Common Objects in Context.
\newblock arXiv:1405.0312.

\bibitem[{Liu et~al.(2016)Liu, Anguelov, Erhan, Szegedy, Reed, Fu, and
  Berg}]{ssd}
Liu, W.; Anguelov, D.; Erhan, D.; Szegedy, C.; Reed, S.; Fu, C.-Y.; and Berg,
  A.~C. 2016.
\newblock SSD: Single Shot MultiBox Detector.
\newblock \emph{Lecture Notes in Computer Science}, 21–37.

\bibitem[{Loshchilov and Hutter(2017)}]{cosan}
Loshchilov, I.; and Hutter, F. 2017.
\newblock SGDR: Stochastic Gradient Descent with Warm Restarts.
\newblock arXiv:1608.03983.

\bibitem[{Ma et~al.(2021)Ma, Zhang, Liu, and Sun}]{acon}
Ma, N.; Zhang, X.; Liu, M.; and Sun, J. 2021.
\newblock Activate or Not: Learning Customized Activation.
\newblock arXiv:2009.04759.

\bibitem[{Ma et~al.(2018)Ma, Zhang, Zheng, and Sun}]{shufflenet}
Ma, N.; Zhang, X.; Zheng, H.-T.; and Sun, J. 2018.
\newblock ShuffleNet V2: Practical Guidelines for Efficient CNN Architecture
  Design.
\newblock arXiv:1807.11164.

\bibitem[{Maas, Hannun, and Ng(2013)}]{lrelu}
Maas, A.~L.; Hannun, A.~Y.; and Ng, A.~Y. 2013.
\newblock Rectifier nonlinearities improve neural network acoustic models.
\newblock In \emph{in ICML Workshop on Deep Learning for Audio, Speech and
  Language Processing}.

\bibitem[{Misra(2020)}]{mish}
Misra, D. 2020.
\newblock Mish: A Self Regularized Non-Monotonic Activation Function.
\newblock arXiv:1908.08681.

\bibitem[{Molina, Schramowski, and Kersting(2020)}]{pau}
Molina, A.; Schramowski, P.; and Kersting, K. 2020.
\newblock Pad\'e Activation Units: End-to-end Learning of Flexible Activation
  Functions in Deep Networks.
\newblock arXiv:1907.06732.

\bibitem[{Nag and Bhattacharyya(2021)}]{serf}
Nag, S.; and Bhattacharyya, M. 2021.
\newblock SERF: Towards better training of deep neural networks using
  log-Softplus ERror activation Function.
\newblock arXiv:2108.09598.

\bibitem[{Nair and Hinton(2010)}]{relu}
Nair, V.; and Hinton, G.~E. 2010.
\newblock Rectified Linear Units Improve Restricted Boltzmann Machines.
\newblock In F{\"{u}}rnkranz, J.; and Joachims, T., eds., \emph{Proceedings of
  the 27th International Conference on Machine Learning (ICML-10), June 21-24,
  2010, Haifa, Israel}, 807--814. Omnipress.

\bibitem[{Netzer et~al.(2011)Netzer, Wang, Coates, Bissacco, Wu, and Ng}]{SVHN}
Netzer, Y.; Wang, T.; Coates, A.; Bissacco, A.; Wu, B.; and Ng, A.~Y. 2011.
\newblock Reading Digits in Natural Images with Unsupervised Feature Learning.

\bibitem[{Radford et~al.(2019)Radford, Wu, Child, Luan, Amodei, and
  Sutskever}]{gpt2}
Radford, A.; Wu, J.; Child, R.; Luan, D.; Amodei, D.; and Sutskever, I. 2019.
\newblock {Language Models are Unsupervised Multitask Learners}.

\bibitem[{Ramachandran, Zoph, and Le(2017)}]{swish}
Ramachandran, P.; Zoph, B.; and Le, Q.~V. 2017.
\newblock Searching for Activation Functions.
\newblock arXiv:1710.05941.

\bibitem[{Robbins and Monro(1951)}]{sgd1}
Robbins, H.; and Monro, S. 1951.
\newblock A stochastic approximation method.
\newblock \emph{Annals of Mathematical Statistics}, 22: 400--407.

\bibitem[{Ronneberger, Fischer, and Brox(2015)}]{unet}
Ronneberger, O.; Fischer, P.; and Brox, T. 2015.
\newblock U-Net: Convolutional Networks for Biomedical Image Segmentation.
\newblock arXiv:1505.04597.

\bibitem[{Sandler et~al.(2019)Sandler, Howard, Zhu, Zhmoginov, and
  Chen}]{mobile}
Sandler, M.; Howard, A.; Zhu, M.; Zhmoginov, A.; and Chen, L.-C. 2019.
\newblock MobileNetV2: Inverted Residuals and Linear Bottlenecks.
\newblock arXiv:1801.04381.

\bibitem[{Simonyan and Zisserman(2015)}]{vgg}
Simonyan, K.; and Zisserman, A. 2015.
\newblock Very Deep Convolutional Networks for Large-Scale Image Recognition.
\newblock arXiv:1409.1556.

\bibitem[{Srivastava et~al.(2014)Srivastava, Hinton, Krizhevsky, Sutskever, and
  Salakhutdinov}]{dropout}
Srivastava, N.; Hinton, G.; Krizhevsky, A.; Sutskever, I.; and Salakhutdinov,
  R. 2014.
\newblock Dropout: A Simple Way to Prevent Neural Networks from Overfitting.
\newblock \emph{J. Mach. Learn. Res.}, 15(1): 1929–1958.

\bibitem[{Szegedy et~al.(2014)Szegedy, Liu, Jia, Sermanet, Reed, Anguelov,
  Erhan, Vanhoucke, and Rabinovich}]{googlenet}
Szegedy, C.; Liu, W.; Jia, Y.; Sermanet, P.; Reed, S.; Anguelov, D.; Erhan, D.;
  Vanhoucke, V.; and Rabinovich, A. 2014.
\newblock Going Deeper with Convolutions.
\newblock arXiv:1409.4842.

\bibitem[{Szegedy et~al.(2015)Szegedy, Vanhoucke, Ioffe, Shlens, and
  Wojna}]{incep}
Szegedy, C.; Vanhoucke, V.; Ioffe, S.; Shlens, J.; and Wojna, Z. 2015.
\newblock Rethinking the Inception Architecture for Computer Vision.
\newblock arXiv:1512.00567.

\bibitem[{Tan and Le(2020)}]{efficientnet}
Tan, M.; and Le, Q.~V. 2020.
\newblock EfficientNet: Rethinking Model Scaling for Convolutional Neural
  Networks.
\newblock arXiv:1905.11946.

\bibitem[{Vaswani et~al.(2017)Vaswani, Shazeer, Parmar, Uszkoreit, Jones,
  Gomez, Kaiser, and Polosukhin}]{attn}
Vaswani, A.; Shazeer, N.; Parmar, N.; Uszkoreit, J.; Jones, L.; Gomez, A.~N.;
  Kaiser, L.; and Polosukhin, I. 2017.
\newblock Attention Is All You Need.
\newblock arXiv:1706.03762.

\bibitem[{Xiao, Rasul, and Vollgraf(2017)}]{fashion}
Xiao, H.; Rasul, K.; and Vollgraf, R. 2017.
\newblock Fashion-mnist: a novel image dataset for benchmarking machine
  learning algorithms.
\newblock \emph{arXiv preprint arXiv:1708.07747}.

\bibitem[{Xie et~al.(2017)Xie, Girshick, Dollár, Tu, and He}]{resnext}
Xie, S.; Girshick, R.; Dollár, P.; Tu, Z.; and He, K. 2017.
\newblock Aggregated Residual Transformations for Deep Neural Networks.
\newblock arXiv:1611.05431.

\bibitem[{Yu et~al.(2019)Yu, Wang, Shelhamer, and Darrell}]{dla}
Yu, F.; Wang, D.; Shelhamer, E.; and Darrell, T. 2019.
\newblock Deep Layer Aggregation.
\newblock arXiv:1707.06484.

\bibitem[{Zagoruyko and Komodakis(2016)}]{wrn}
Zagoruyko, S.; and Komodakis, N. 2016.
\newblock Wide Residual Networks.
\newblock arXiv:1605.07146.

\bibitem[{Zhang et~al.(2018)Zhang, Cisse, Dauphin, and Lopez-Paz}]{mixup}
Zhang, H.; Cisse, M.; Dauphin, Y.~N.; and Lopez-Paz, D. 2018.
\newblock mixup: Beyond Empirical Risk Minimization.
\newblock In \emph{International Conference on Learning Representations}.

\bibitem[{Zheng et~al.(2015)Zheng, Yang, Liu, Liang, and Li}]{softplus}
Zheng, H.; Yang, Z.; Liu, W.; Liang, J.; and Li, Y. 2015.
\newblock Improving deep neural networks using softplus units.
\newblock In \emph{2015 International Joint Conference on Neural Networks
  (IJCNN)}, 1--4.

\end{thebibliography}

\newpage
\appendix
\section{Supplementary Material}

\section{Baseline activation functions}
\begin{enumerate}
    \item \textbf{ReLU:} ReLU \cite{relu} is one of the most widely used activation function and defined as $f(x) = max(x,0)$.
    \item \textbf{Leaky ReLU:} Leaky ReLU \cite{lrelu} is a variant of ReLU in which a linear negative component is added to ReLU. Leaky ReLU is defined as $f(x) = max(x,0.01 x)$
    \item  \textbf{Parametric ReLU:} Parametric ReLU (PReLU) \cite{prelu} is a trainable version of Leaky ReLU. PReLU is defined as $f(x) = max(x,\alpha x)$ where $\alpha$ is the learnable parameter.
    \item \textbf{ReLU6:} ReLU6 \cite{relu6} is defined as $f(x) = min(max(x, 0), 6)$.
    \item \textbf{ELU:} ELU \cite{elu} is defined as $f(x) = max(x,0) + min(\alpha(e^x - 1), 0)$.
    \item \textbf{Softplus:} Softplus \cite{softplus} is a smoothing of ReLU. SoftPlus is defined as $f(x) = ln(1+e^x)$.
    \item \textbf{Swish:} Swish \cite{swish} is a trainable activation function found by automated search technique by Google bain team. Swish is defined as $f(x) = \dfrac{x}{1+e^{-\beta x}}$, where $\beta$ is a trainable parameter.
    \item \textbf{GELU:} GELU \cite{gelu} can be viewed as a smoothing version of ReLU. GELU is defined as $f(x) = \frac{x}{2}(1+erf(\frac{x}{\sqrt{2}})$ or $f(x) = 0.5x(1 + tanh[\sqrt{\dfrac{2}{\pi}}(x + 0.044715x^3)])$.

     \item \textbf{Mish:} Mish \cite{mish} is defined as $x\text{tanh}(ln(1+e^x))$.
     
    \item \textbf{Pad\'e Activation Unit (PAU):} PAU \cite{pau} is a popular activation function, and it is defined as the rational approximation the of Leaky ReLU function.
\end{enumerate}


\section{Experimental results on MNIST, Fashion MNIST, and SVHN datasets}
In this supplementary section, we present more experiments on MNIST, Fashion MNIST, and SVHN datasets with LeNet \cite{lenet} and an 8-layer homogeneous custom convolutional neural network (CNN) architecture and the results are reported on Table~\ref{tabl22} and \ref{tabll22}. The custom network is constructed with CNN layers with $3\times 3$ kernels and max-pooling layers with $2\times 2$ kernels. We have used Channel depths of size 128 (twice), 64 (thrice), 32 (twice), with a dense layer of size 128, Max-pooling layer(thrice), and dropout\cite{dropout}. We have applied batch-normalization\cite{batch} before the activation function layer.\\
\section{Experimental results on CIFAR datasets}
A more detailed experiments on CIFAR10 and CIFAR100 datasets with EfficientNet B0 (EN-B0) \cite{efficientnet}, LeNet (LN) \cite{lenet}, AlexNet (AN) \cite{alexnet}, PreActResnet-18 (PARN-18) \cite{preactresnet}, Deep Layer Aggregation (DLA) \cite{dla}, Googlenet (GN) \cite{googlenet}, Resnext-50 (Rxt) \cite{resnext}, Xception (Xpt) \cite{xception}, ShuffleNet V2 (SN-V2) \cite{shufflenet}, ResNet18 (RN-18) \cite{resnet}, and Network in Network (NIN) \cite{nin} is reported in the Table~\ref{tabe2} and ~\ref{tabee2} respectively. We get good improvement with EfficientNet B0, PreActResnet-18, LeNet, GoogleNet, Resnext-50, and ShuffleNet V2 models on both the datasets compared to ReLU or other baseline activation functions. 

We report more detailed results with Mixup \cite{mixup} augmentation method with ShuffleNet V2 (2.0x) and ResNet-18 models in Table\ref{tabmixup}. The table shows that the proposed activations beat the baseline activation functions in both models with Mixup augmentation. We consider the same experimental setup for Mixup as reported in the CIFAR section.
\section{ImageNet-1k}
ImageNet-1k \cite{imagenet} is a widely used computer vision database with more than 1.2 million training images and have 1000 different classes. We report result with ShuffleNet V2 (1.0x) \cite{shufflenet} model on ImageNet-1k dataset in Table~\ref{tabin}. We use SGD optimizer (\cite{sgd1}, \cite{sgd2}), 0.9 momentum, $5e^{-4}$ weight decay, and a batch size of 256 and trained upto 600k iterations. Experiments on ImageNet-1k is conducted on four NVIDIA V100 GPUs with 32GB RAM each.

\begin{table}[!htbp]
\begin{center}
\begin{tabular}{ |c|c|c|c| }
 \hline
 Activation Function & ShuffleNet V2 (1.0x) \\
  
 \hline
 ReLU  & 69.20 \\ 
  \hline
 Leaky ReLU & 69.32 \\
 \hline
  
 PReLU & 69.28 \\
 \hline
 ReLU6 & 69.40 \\

 \hline

 ELU  & 69.24 \\
 \hline
 Softplus & 69.07 \\
 \hline
  Swish  & 70.06\\
  
\hline
 GELU & 69.91 \\
  \hline
  Mish  & 69.95\\
 \hline
 PAU & 70.17 \\
 \hline
 ErfAct & \textbf{70.65} \\ 
  \hline
 Pserf & \textbf{70.57} \\ 
\hline
 \end{tabular}
 \caption{top-1 Accuracy reported on ImageNet-1k dataset.} 
\label{tabin}
\end{center}

\end{table}

\begin{table*}[!htbp]
\begin{center}
\begin{tabular}{ |c|c|c|c|c| }
 \hline
 Activation Function &  \makecell{MNIST} & \makecell{Fashion MNIST} & \makecell{SVHN}  \\
 \hline
 ReLU  &  99.09$\pm$0.10 & 92.92$\pm$0.21  & 95.10$\pm$0.22\\ 
 \hline
  Swish  & 99.20$\pm$0.09  & 93.04$\pm$0.23 & 95.21$\pm$0.23\\
 \hline
 
 Leaky ReLU($\alpha$ = 0.01) & 99.14$\pm$0.09  & 92.99$\pm$0.22 & 95.30$\pm$0.25\\
 \hline
  ELU  & 99.10$\pm$0.13 & 92.91$\pm$0.30 & 95.17$\pm$0.27\\
 \hline
 Softplus & 98.95 $\pm$0.17 & 92.72$\pm$0.28 & 95.08$\pm$0.37\\
 \hline
 Mish & 99.32$\pm$0.10 & 93.12$\pm$0.21 & 95.33$\pm$0.21\\
 \hline
 
 GELU & 99.28$\pm$0.09 & 93.19$\pm$0.22 & 95.22$\pm$0.24\\
 \hline
 PReLU & 99.08$\pm$0.17  & 92.89$\pm$0.35 & 95.15$\pm$0.30\\
 \hline
 ReLU6 & 99.17$\pm$0.12 & 92.99$\pm$0.20 & 95.17$\pm$0.22\\
  \hline
 
 PAU & 99.24$\pm$0.10 & 93.24$\pm$0.20 & 95.15$\pm$0.23\\
 \hline
 ErfAct & \textbf{99.42}$\pm$0.08 & \textbf{93.42}$\pm$0.23 & \textbf{95.49}$\pm$0.24\\
 \hline
 Pserf& \textbf{99.40}$\pm$0.08  & \textbf{93.35}$\pm$0.20 & \textbf{95.55}$\pm$0.23\\
 \hline

 \end{tabular}
\caption{Comparison between different baseline activations and ErfAct and Pserf on MNIST, Fashion MNIST, and SVHN datasets with Custom designed network. 10-fold mean accuracy (in \%) have been reported. mean$\pm$std is reported in the table.} 
\label{tabl22}
\end{center}
\end{table*}

\begin{table*}[!htbp]
\begin{center}
\begin{tabular}{ |c|c|c|c|c| }
 \hline
 Activation Function &  \makecell{MNIST} & \makecell{Fashion MNIST} & \makecell{SVHN}  \\
 \hline
 ReLU  &  98.95$\pm$0.11 & 91.00$\pm$0.20  & 93.17$\pm$0.24\\ 
 \hline
  Swish  & 99.04$\pm$0.11  & 91.15$\pm$0.22 & 93.22$\pm$0.21\\
 \hline
 
 Leaky ReLU($\alpha$ = 0.01) & 99.02$\pm$0.10  & 91.05$\pm$0.20 & 93.25$\pm$0.24\\
 \hline
  ELU  & 98.95$\pm$0.12 & 91.98$\pm$0.28 & 93.11$\pm$0.24\\
 \hline
 Softplus & 98.81 $\pm$0.14 & 90.81$\pm$0.29 & 93.08$\pm$0.37\\
 \hline
 Mish & 99.12$\pm$0.11 & 91.12$\pm$0.21 & 93.30$\pm$0.21\\
 \hline
 
 GELU & 99.15$\pm$0.10 & 91.17$\pm$0.20 & 93.22$\pm$0.21\\
 \hline
 PReLU & 99.01$\pm$0.17  & 90.89$\pm$0.25 & 93.05$\pm$0.28\\
 \hline
 ReLU6 & 99.07$\pm$0.10 & 90.99$\pm$0.24 & 93.10$\pm$0.20\\
  \hline
 
 PAU & 99.14$\pm$0.09 & 91.20$\pm$0.19 & 93.17$\pm$0.20\\
 \hline
 ErfAct & \textbf{99.30}$\pm$0.08 & \textbf{91.37}$\pm$0.20 & \textbf{93.52}$\pm$0.21\\
 \hline
 Pserf& \textbf{99.32}$\pm$0.08  & \textbf{91.31}$\pm$0.22 & \textbf{93.43}$\pm$0.21\\
 \hline

 \end{tabular}
\caption{Comparison between different baseline activations and ErfAct and Pserf on MNIST, Fashion MNIST, and SVHN datasets with LeNet model. 10-fold mean accuracy (in \%) have been reported. mean$\pm$std is reported in the table.} 
\label{tabll22}
\end{center}
\end{table*}
\begin{table*}[!htbp]

\begin{center}
\begin{tabular}{ |c|c|c|c| }
 \hline
 Activation Function & ShuffleNet V2 (2.0x) &   ResNet 18 \\
  
 \hline
 ReLU  & 70.02 $\pm$ 0.22 &  73.72 $\pm$ 0.23\\

  \hline
  Leaky ReLU & 69.85 $\pm$ 0.24  & 73.91 $\pm$ 0.24\\

 \hline

 ELU  & 70.25 $\pm$ 0.23  & 73.92 $\pm$ 0.26\\
  \hline
Swish  & 73.12 $\pm$ 0.23  & 74.52 $\pm$ 0.23\\
  \hline
 Softplus & 69.52 $\pm$ 0.30  & 73.63 $\pm$ 0.26\\
 
 \hline
 Mish  & 73.65 $\pm$ 0.21  & 74.97 $\pm$ 0.24\\
 \hline

 GELU & 73.25 $\pm$ 0.23  & 74.45 $\pm$ 0.23 \\
 \hline
  
 PReLU & 70.05 $\pm$ 0.23  & 74.10 $\pm$ 0.27\\

\hline
 ReLU6 & 70.20 $\pm$ 0.24  & 74.01 $\pm$ 0.24\\

 \hline
 PAU & 73.28 $\pm$ 0.24  & 74.65 $\pm$ 0.25\\
 \hline
 ErfAct & \textbf{75.07} $\pm$ 0.22   &  \textbf{75.67} $\pm$ 0.21 \\ 
 \hline
 Pserf & \textbf{74.84} $\pm$ 0.24   &  \textbf{75.46} $\pm$ 0.21 \\
\hline
 \end{tabular}
\caption{Comparison between different baseline activations and ErfAct and Pserf on CIFAR100 dataset. Top-1 accuracy(in $\%$) with Mixup augmentation method for mean of 12 different runs have been reported. mean$\pm$std is reported in the table.} 
\label{tabmixup}
\end{center}
\end{table*}



\begin{table*}[!htbp]
\footnotesize
\begin{center}
\begin{tabular}{ |c|c|c|c|c|c|c|c|c|c|c|c|c|c| }
 \hline
\makecell{Activation\\ Function} & EN-B0 & LN & AN & PARN-18 & DLA  & GN & Rxt & Xpt & SN-V1 & RN-18 & NIN\\

 \hline 
 ReLU & \makecell{95.04\\$\pm$0.16} & \makecell{75.68\\$\pm$0.21} &   \makecell{84.18\\$\pm$0.21} &  \makecell{93.47\\$\pm$0.22} & \makecell{93.90\\$\pm$0.18} & \makecell{93.02\\$\pm$0.20} & \makecell{93.28\\$\pm$0.18} & \makecell{90.64\\$\pm$0.22} & \makecell{94.20\\$\pm$0.20} & \makecell{94.01\\$\pm$0.21} & \makecell{90.49\\$\pm$0.24}\\ 
 \hline
 \makecell{Leaky ReLU \\($\alpha$ = 0.01)} & \makecell{95.22\\$\pm$0.16} & \makecell{75.91\\$\pm$0.22} &   \makecell{84.32\\$\pm$0.23} &  \makecell{93.61\\$\pm$0.21} & \makecell{94.01\\$\pm$0.20} & \makecell{92.91\\$\pm$0.18} & \makecell{93.39\\$\pm$0.19} & \makecell{90.80\\$\pm$0.24} & \makecell{94.32\\$\pm$0.22} & \makecell{94.12\\$\pm$0.24} & \makecell{90.59\\$\pm$0.26}\\ 
 \hline
 ELU & \makecell{95.35\\$\pm$0.18} & \makecell{76.10\\$\pm$0.20} &   \makecell{84.78\\$\pm$0.20} &  \makecell{93.65\\$\pm$0.22} & \makecell{93.96\\$\pm$0.20} & \makecell{93.06\\$\pm$0.17} & \makecell{93.55\\$\pm$0.23} & \makecell{91.38\\$\pm$0.24} & \makecell{94.32\\$\pm$0.21} & \makecell{94.19\\$\pm$0.24} & \makecell{90.55\\$\pm$0.24}\\ 
 \hline
 Swish & \makecell{95.60\\$\pm$0.17} & \makecell{77.55\\$\pm$0.19} &   \makecell{85.10\\$\pm$0.20} &  \makecell{93.87\\$\pm$0.20} & \makecell{94.25\\$\pm$0.17} & \makecell{93.30\\$\pm$0.20} & \makecell{93.69\\$\pm$0.19} & \makecell{91.94\\$\pm$0.20} & \makecell{94.65\\$\pm$0.19} & \makecell{94.29\\$\pm$0.21} & \makecell{90.97\\$\pm$0.25}\\ 
 \hline
 Softplus & \makecell{95.10\\$\pm$0.27} & \makecell{75.65\\$\pm$0.33} &   \makecell{84.22\\$\pm$0.30} &  \makecell{93.12\\$\pm$0.26} & \makecell{93.71\\$\pm$0.25} & \makecell{92.64\\$\pm$0.29} & \makecell{93.01\\$\pm$0.29} & \makecell{90.69\\$\pm$0.30} & \makecell{93.92\\$\pm$0.25} & \makecell{93.99\\$\pm$0.27} & \makecell{90.39\\$\pm$0.30}\\
 \hline
 Mish & \makecell{95.75\\$\pm$0.15} & \makecell{\textbf{78.76}\\$\pm$0.16} &   \makecell{85.70\\$\pm$0.18} &  \makecell{93.70\\$\pm$0.24} & \makecell{94.40\\$\pm$0.17} & \makecell{93.22\\$\pm$0.20} & \makecell{93.92\\$\pm$0.17} & \makecell{92.15\\$\pm$0.19} & \makecell{94.78\\$\pm$0.19} & \makecell{94.45\\$\pm$0.25} & \makecell{\textbf{91.17}\\$\pm$0.23}\\ 
 \hline
 GELU & \makecell{95.39\\$\pm$0.19} & \makecell{77.79\\$\pm$0.17} &   \makecell{85.15\\$\pm$0.21} &  \makecell{93.77\\$\pm$0.19} & \makecell{94.12\\$\pm$0.20} & \makecell{93.45\\$\pm$0.20} & \makecell{93.77\\$\pm$0.19} & \makecell{91.89\\$\pm$0.22} & \makecell{94.55\\$\pm$0.21} & \makecell{94.45\\$\pm$0.24} & \makecell{91.01\\$\pm$0.24}\\ 
 \hline
 PReLU & \makecell{95.20\\$\pm$0.18} & \makecell{75.85\\$\pm$0.24} &   \makecell{84.38\\$\pm$0.23} &  \makecell{93.46\\$\pm$0.25} & \makecell{93.01\\$\pm$0.22} & \makecell{92.89\\$\pm$0.24} & \makecell{93.45\\$\pm$0.26} & \makecell{91.29\\$\pm$0.26} & \makecell{94.34\\$\pm$0.22}
 & \makecell{94.15\\$\pm$0.28}
 & \makecell{90.83\\$\pm$0.27}\\
 \hline
 ReLU6 & \makecell{95.43\\$\pm$0.16} & \makecell{75.71\\$\pm$0.18} &   \makecell{84.64\\$\pm$0.22} &  \makecell{93.75\\$\pm$0.22} & \makecell{94.09\\$\pm$0.17} & \makecell{92.89\\$\pm$0.18} & \makecell{93.48\\$\pm$0.22} & \makecell{91.38\\$\pm$0.24} & \makecell{94.22\\$\pm$0.20}
 & \makecell{94.28\\$\pm$0.24}
 & \makecell{90.87\\$\pm$0.24}\\
 \hline
 PAU & \makecell{95.35\\$\pm$0.17} & \makecell{77.59\\$\pm$0.21} &   \makecell{85.01\\$\pm$0.24} &  \makecell{93.75\\$\pm$0.22} & \makecell{94.34\\$\pm$0.20} & \makecell{93.29\\$\pm$0.21} & \makecell{93.52\\$\pm$0.20} & \makecell{91.79\\$\pm$0.21} & \makecell{94.60\\$\pm$0.21}
 & \makecell{94.31\\$\pm$0.21}
 & \makecell{90.97\\$\pm$0.25}\\
 \hline
 ErfAct & \makecell{\textbf{96.10}\\$\pm$0.15} & \makecell{77.48\\$\pm$0.19} &   \makecell{\textbf{87.01}\\$\pm$0.20} &  \makecell{\textbf{94.10}\\$\pm$0.20} & \makecell{\textbf{94.67}\\$\pm$0.17} & \makecell{\textbf{94.12}\\$\pm$0.18} & \makecell{\textbf{94.17}\\$\pm$0.18} & \makecell{\textbf{93.01}\\$\pm$0.17} & \makecell{\textbf{95.14}\\$\pm$0.19}
 & \makecell{\textbf{94.71}\\$\pm$0.23}
 & \makecell{90.81\\$\pm$0.24}\\ 
 \hline
 Pserf & \makecell{\textbf{95.98}\\$\pm$0.18} & \makecell{77.52\\$\pm$0.20} &   \makecell{\textbf{87.15}\\$\pm$0.20} &  \makecell{\textbf{94.01}\\$\pm$0.22} & \makecell{\textbf{94.56}\\$\pm$0.20} & \makecell{\textbf{93.95}\\$\pm$0.20} & \makecell{\textbf{94.01}\\$\pm$0.19} & \makecell{\textbf{93.18}\\$\pm$0.18} & \makecell{\textbf{94.96}\\$\pm$0.18}
 & \makecell{\textbf{94.68}\\$\pm$0.18}
 & \makecell{90.78\\$\pm$0.24}\\
 \hline
  
 \end{tabular}
\caption{Comparison between different baseline activations and ErfAct and Pserf on CIFAR10 dataset. Top-1 accuracy(in $\%$) for mean of 12 different runs have been reported. mean$\pm$std is reported in the table.} 
\label{tabe2}
\end{center}
\end{table*}

\begin{table*}[!htbp]
\footnotesize
\begin{center}
\begin{tabular}{ |c|c|c|c|c|c|c|c|c|c|c|c|c|c| }
 \hline
\makecell{Activation\\ Function} & EN-B0 & LN & AN & PARN-18 & DLA  & GN & Rxt & Xpt & SN-V1 & RN-18 & NIN\\

 \hline 
 ReLU & \makecell{76.45\\$\pm$0.26} & \makecell{45.50\\$\pm$0.30} &   \makecell{55.02\\$\pm$0.30} &  \makecell{73.10\\$\pm$0.20} & \makecell{74.50\\$\pm$0.22} & \makecell{72.64\\$\pm$0.28} & \makecell{74.31\\$\pm$0.22} & \makecell{71.20\\$\pm$0.20} & \makecell{73.70\\$\pm$0.23} & \makecell{73.17\\$\pm$0.25} & \makecell{65.12\\$\pm$0.25}\\ 
 \hline
 \makecell{Leaky ReLU \\($\alpha$ = 0.01)} & \makecell{76.70\\$\pm$0.25} & \makecell{45.64\\$\pm$0.28} &   \makecell{55.34\\$\pm$0.28} &  \makecell{73.30\\$\pm$0.21} & \makecell{74.62\\$\pm$0.23} & \makecell{72.51\\$\pm$0.28} & \makecell{74.60\\$\pm$0.23} & \makecell{71.10\\$\pm$0.24} & \makecell{73.89\\$\pm$0.25} & \makecell{73.21\\$\pm$0.23} & \makecell{65.27\\$\pm$0.23}\\
 \hline
 ELU & \makecell{76.77\\$\pm$0.26} & \makecell{45.23\\$\pm$0.27} &   \makecell{55.72\\$\pm$0.28} &  \makecell{73.41\\$\pm$0.23} & \makecell{74.54\\$\pm$0.24} & \makecell{72.85\\$\pm$0.27} & \makecell{74.71\\$\pm$0.24} & \makecell{71.40\\$\pm$0.22} & \makecell{73.98\\$\pm$0.21} & \makecell{73.40\\$\pm$0.25} & \makecell{65.39\\$\pm$0.23}\\
 \hline
 Swish & \makecell{77.34\\$\pm$0.20} & \makecell{47.30\\$\pm$0.25} &   \makecell{57.64\\$\pm$0.28} &  \makecell{74.98\\$\pm$0.24} & \makecell{75.20\\$\pm$0.20} & \makecell{73.45\\$\pm$0.28} & \makecell{75.06\\$\pm$0.26} & \makecell{72.16\\$\pm$0.24} & \makecell{74.29\\$\pm$0.22} & \makecell{73.65\\$\pm$0.24} & \makecell{66.20\\$\pm$0.22}\\
 \hline
 Softplus & \makecell{76.41\\$\pm$0.30} & \makecell{44.10\\$\pm$0.38} &   \makecell{54.85\\$\pm$0.36} &  \makecell{73.10\\$\pm$0.35} & \makecell{74.31\\$\pm$0.26} & \makecell{72.09\\$\pm$0.35} & \makecell{74.20\\$\pm$0.34} & \makecell{71.51\\$\pm$0.36} & \makecell{73.90\\$\pm$0.27} & \makecell{72.80\\$\pm$0.36} & \makecell{65.25\\$\pm$0.30}\\
 \hline
 Mish & \makecell{78.02\\$\pm$0.23} & \makecell{\textbf{47.49}\\$\pm$0.28} &   \makecell{58.35\\$\pm$0.25} &  \makecell{74.84\\$\pm$0.24} & \makecell{75.45\\$\pm$0.20} & \makecell{73.85\\$\pm$0.25} & \makecell{76.07\\$\pm$0.24} & \makecell{73.34\\$\pm$0.23} & \makecell{74.40\\$\pm$0.21} & \makecell{74.39\\$\pm$0.22} & \makecell{\textbf{66.50}\\$\pm$0.22}\\
 \hline
 GELU & \makecell{77.30\\$\pm$0.24} & \makecell{47.23\\$\pm$0.25} &   \makecell{57.55\\$\pm$0.27} &  \makecell{74.87\\$\pm$0.23} & \makecell{75.20\\$\pm$0.23} & \makecell{73.32\\$\pm$0.27} & \makecell{75.32\\$\pm$0.23} & \makecell{72.25\\$\pm$0.22} & \makecell{73.15\\$\pm$0.22} & \makecell{73.77\\$\pm$0.22} & \makecell{66.01\\$\pm$0.22}\\
 \hline
 PReLU & \makecell{76.62\\$\pm$0.28} & \makecell{45.69\\$\pm$0.30} &   \makecell{55.41\\$\pm$0.30} &  \makecell{73.16\\$\pm$0.25} & \makecell{74.98\\$\pm$0.24} & \makecell{72.60\\$\pm$0.30} & \makecell{74.50\\$\pm$0.26} & \makecell{71.30\\$\pm$0.23} & \makecell{73.79\\$\pm$0.24} & \makecell{73.10\\$\pm$0.26} & \makecell{65.56\\$\pm$0.27}\\
 \hline
 ReLU6 & \makecell{76.58\\$\pm$0.23} & \makecell{45.86\\$\pm$0.28} &   \makecell{55.75\\$\pm$0.28} &  \makecell{73.30\\$\pm$0.25} & \makecell{74.69\\$\pm$0.21} & \makecell{72.40\\$\pm$0.24} & \makecell{74.69\\$\pm$0.24} & \makecell{71.40\\$\pm$0.24} & \makecell{73.99\\$\pm$0.23} & \makecell{73.30\\$\pm$0.25} & \makecell{65.42\\$\pm$0.24}\\
 \hline
 PAU & \makecell{77.21\\$\pm$0.26} & \makecell{47.17\\$\pm$0.28} &   \makecell{57.42\\$\pm$0.27} &  \makecell{74.71\\$\pm$0.22} & \makecell{75.50\\$\pm$0.22} & \makecell{73.60\\$\pm$0.28} & \makecell{75.60\\$\pm$0.25} & \makecell{72.50\\$\pm$0.24} & \makecell{74.36\\$\pm$0.22} & \makecell{73.99\\$\pm$0.22} & \makecell{66.20\\$\pm$0.22}\\
 \hline
 ErfAct & \makecell{\textbf{78.97}\\$\pm$0.23} & \makecell{47.29\\$\pm$0.26} &   \makecell{\textbf{60.89}\\$\pm$0.25} &  \makecell{\textbf{75.77}\\$\pm$0.24} & \makecell{\textbf{76.43}\\$\pm$0.18} & \makecell{\textbf{74.47}\\$\pm$0.26} & \makecell{\textbf{77.23}\\$\pm$0.23} & \makecell{\textbf{74.32}\\$\pm$0.22} & \makecell{\textbf{74.90}\\$\pm$0.21} & \makecell{\textbf{74.79}\\$\pm$0.24} & \makecell{66.25\\$\pm$0.22}\\ 
 \hline
 Pserf & \makecell{\textbf{78.75}\\$\pm$0.24} & \makecell{47.27\\$\pm$0.27} &   \makecell{\textbf{60.57}\\$\pm$0.24} &  \makecell{\textbf{75.60}\\$\pm$0.25} & \makecell{\textbf{76.23}\\$\pm$0.20} & \makecell{\textbf{74.50}\\$\pm$0.25} & \makecell{\textbf{77.10}\\$\pm$0.24} & \makecell{\textbf{74.20}\\$\pm$0.24} & \makecell{\textbf{74.72}\\$\pm$0.20} & \makecell{\textbf{74.84}\\$\pm$0.23} & \makecell{66.35\\$\pm$0.23}\\
 \hline
  
 \end{tabular}
\caption{Comparison between different baseline activations and ErfAct and Pserf on CIFAR100 dataset. Top-1 accuracy(in $\%$) for mean of 12 different runs have been reported. mean$\pm$std is reported in the table.} 
\label{tabee2}
\end{center}
\end{table*}



\end{document}